\documentclass[sigconf, nonacm]{acmart}

\usepackage{algorithm}
\usepackage{algpseudocode}% <========== http://ctan.org/pkg/algorithmicx
\usepackage{diagbox}
\usepackage{multirow}
\usepackage{subfig}
\usepackage{soul}
\usepackage{url}
\usepackage[normalem]{ulem}
\usepackage{pdfpages}

\AtBeginDocument{%
  \providecommand\BibTeX{{%
    \normalfont B\kern-0.5em{\scshape i\kern-0.25em b}\kern-0.8em\TeX}}}

\begin{document}\sloppy

\title{Improving LSHADE by means of a pre-screening mechanism}
%\title{psLSHADE: the LSHADE algorithm enhanced with a pre-screening mechanism}

% \author{Mateusz Zaborski}
% \email{M.Zaborski@mini.pw.edu.pl}
% \orcid{0000-0002-9371-1690}
% \author{Jacek Ma{\'n}dziuk}
% \email{mandziuk@mini.pw.edu.pl}
% \orcid{0000-0003-0947-028X}
% \affiliation{%
%   \institution{Faculty of Mathematics and Information Science, Warsaw University of Technology}
%   \streetaddress{Koszykowa 75}
%   \city{Warsaw}
%   \country{Poland}
%   \postcode{00-662}
% }

% \author{{\color{blue}Mateusz Zaborski and Jacek Ma{\'n}dziuk}}
% \email{{M.Zaborski, mandziuk}@mini.pw.edu.pl}
% \affiliation{%
%   \institution{Faculty of Mathematics and Information Science, Warsaw University of Technology}
%   \streetaddress{Koszykowa 75}
%   \city{Warsaw}
%   \country{Poland}
%   \postcode{00-662}
% }
% \affiliation{%
%   \institution{Faculty of Mathematics and Information Science, Warsaw University of Technology, Warsaw, Poland}
% }

\author{Mateusz Zaborski}
\authornote{Both authors contributed equally to this research.}
\email{M.Zaborski@mini.pw.edu.pl}
\affiliation{%
  \institution{Warsaw University of Technology}
  \streetaddress{Koszykowa 75}
  \city{Warsaw}
  \country{Poland}
  \postcode{00-662}
}

\author{Jacek Ma{\'n}dziuk}
\authornotemark[1]
\email{mandziuk@mini.pw.edu.pl}
\affiliation{%
  \institution{Warsaw University of Technology}
  \streetaddress{Koszykowa 75}
  \city{Warsaw}
  \country{Poland}
  \postcode{00-662}
}

%%%%%%%%%%%%%%%%%%%%%%%%%%%%%%%%%%%%%%%%%%%%%%%%%%%%%%%
%%% ACKNOWLDGMENTS - pozniej -- AFILIACJE - pozniej %%%
%%%%%%%%%%%%%%%%%%%%%%%%%%%%%%%%%%%%%%%%%%%%%%%%%%%%%%%

%\renewcommand{\shortauthors}{Mateusz Zaborski and Jacek Ma{\'n}dziuk}

\begin{abstract}
Evolutionary algorithms have proven to be highly effective in continuous optimization, especially when numerous fitness function evaluations (FFEs) are possible. In certain cases, however, an expensive optimization approach (i.e. with relatively low number of FFEs) must be taken, and such a setting is considered in this work.
The paper introduces an extension to the well-known LSHADE algorithm in the form of a pre-screening mechanism (psLSHADE). The proposed pre-screening relies on the three following components: a specific initial sampling procedure, an archive of samples, and a global linear meta-model of a fitness function that consists of 6 independent transformations of variables. The pre-screening mechanism preliminary assesses the trial vectors and designates the best one of them for further evaluation with the fitness function.
The performance of psLSHADE is evaluated using the CEC2021 benchmark in an expensive scenario with an optimization budget of $10^2-10^4$ FFEs per dimension. We compare psLSHADE with the baseline LSHADE method and the MadDE algorithm. The results indicate that with restricted optimization budgets 
%($10^2-10^3$) 
psLSHADE visibly outperforms both competitive algorithms. In addition, the use of the pre-screening mechanism results in faster population convergence of psLSHADE compared to LSHADE.
\end{abstract}




\maketitle

\section{Introduction}

Evolutionary algorithms (EAs) are applicable to various global optimization problems~\cite{vikhar2016evolutionary}. In this work, we focus on single-objective continuous optimization, in which EAs have proven to be particularly useful~\cite{boussaid2013survey}.

The widely known Differential Evolution (DE) algorithm~\cite{storn1997differential} was initially designed as a relatively simple but efficient heuristic.
Despite unsophisticated structure, it became the precursor of various more advanced and specialized algorithms. 
For instance, Adaptive Differential Evolution with Optional External Archive~\cite{zhang2009jade} (JADE) introduced parameter adaptation. Then,  Success-History Based Parameter Adaptation for Differential Evolution~\cite{tanabe2013success} (SHADE) presented a more efficient parameter adaptation mechanism utilizing an external memory. 
Afterward, the LSHADE~\cite{tanabe2014improving}, that extends SHADE with Linear Population Size Reduction (LPSR), has demonstrated to ameliorate the performance even further. 
Also, a population restart mechanism appeared to be beneficial in certain experimental settings~\cite{tanabe2015tuning}.
Moreover, several other algorithms enhanced certain DE mechanisms, e.g.
AGSK~\cite{mohamed2020evaluating}, LSHADE\_cnEpSin~\cite{awad2017ensemble}, j2020~\cite{brest2020differential}, IMODE~\cite{sallam2020improved}, jDE~\cite{brest2006self}, or SaDE~\cite{qin2008differential}.

Finally, the recently proposed MadDE algorithm~\cite{biswas2021improving}, presented at the \textit{CEC2021 Special Session and Competition on Single Objective Bound Constrained Optimization}, turned out to be superior 
%algorithm outperforming 
to both LSHADE and LSHADE\_cnEpSin.

%It 
%is worth 
%should be noted that 
Besides DE-related methods there is also 
%that not only DE-like algorithms are efficient in global optimization problems. 
%
another family of effective global optimization algorithms 
%is those incorporating 
that refer to the Covariance Matrix Adaptation (CMA) method. These include CMA-ES~\cite{hansen2003reducing}, IPOP-CMA-ES~\cite{auger2005restart}, or KL-BIPOP-CMA-ES~\cite{yamaguchi2017benchmarking}, PSA-CMA-ES~\cite{nishida2018benchmarking}. 

The vast majority of the above-mentioned DE- and CMA-based algorithms are generally effective and the differences between them often boil down to the details of the experiment scenario, e.g. the utilized benchmarks or method parameterization.

%Nevertheless, when looking at optimization, there is a specific polarity. 
Predominantly, the performance of an optimization algorithm is measured with respect to a number of (true) fitness function evaluations (FFEs).
In many settings, numerous FFEs are assumed while evaluating the algorithm's performance (e.g. the CEC2021 benchmark assumes $2 \cdot 10^5$ FFEs for 10D problems). In contrast, expensive optimization setups are limited to a relatively small number of evaluations (e.g. CEC2015 benchmark for Computationally Expensive Numerical Optimization assumes $5 \cdot 10^2$ FFEs for 10D problems).
Complex surrogate models can be highly supportive in such costly optimization, but their high computational complexity makes them scale poorly (e.g. Efficient Global Optimization~\cite{jones1998efficient}, Kriging~\cite{cressie1990origins}, DTS-CMA-ES~\cite{bajer2019gaussian}).
% {\color{red}As a result, there is a considerable gap between these two optimization scenarios.
% In general, the assumption of a pre-imposed optimization budget, understood as FFEs, can be misleading when evaluating algorithms.}

Surrogate-assisted optimization has received considerable 
%increasing attraction 
attention in recent years~\cite{jin2011surrogate}. The LS-CMA-ES~\cite{auger2004ls} utilizes quadratic approximation of the fitness function. Also, polynomials can be utilized as local meta-models~\cite{kern2006local}.
M-GAPSO~\cite{zaborski2020analysis} is a hybrid of PSO, DE, and local meta-models: linear and quadratic.
SHADE-LM~\cite{okulewicz2021benchmarking} is a new hybrid of SHADE and efficient, in terms of complexity,  local meta-models.
The lq-CMA-ES~\cite{hansen2019global} incorporates a quadratic meta-model to replace fitness evaluations with surrogate estimates to reduce unnecessary FFEs. The replacement depends on the rank correlation measure (Kendall's $\tau$~\cite{kendall1938new}).

This work presents and evaluates psLSHADE: the LSHADE algorithm extended with a pre-screening mechanism. The pre-screening mechanism includes a modified initial sampling procedure, an archive of samples, and a global meta-model of a fitness function. The meta-model is a linear combination
%and consists 
of 6 components (independent variable transformations). It ranks trial vectors and designates the best one of them for further FFE.

% Parallelization (noguchi2021parallel) is a popular method.
The remainder of the paper is structured as follows. Section~\ref{sec:lshade} describes the baseline LSHADE. Section~\ref{sec:pslshade} introduces the principles of psLSHADE, in particular the pre-screening mechanism. Section~\ref{sec:experiment} presents experimental evaluation of psLSHADE using the CEC2021 benchmark in expensive scenarios. Section~\ref{sec:analysis} analyses the meta-model performance. Finally, conclusions and directions for future work are briefly described in Section~\ref{sec:conclusions}.

\section{LSHADE algorithm}\label{sec:lshade}
In this section the LSHADE algorithm is introduced based on its original description~\cite{tanabe2014improving}. 
LSHADE is a single-objective optimization metaheuristic that extends SHADE~\cite{tanabe2013success} with Linear Population Size
Reduction. 
LSHADE includes parameter adaptation, memory, and external archive, similar to the underlying SHADE.
LSHADE is a population-based iterative algorithm where each individual $i$ represents $D$-dimensional point $\pmb{x}^{g}_i = [x^{g}_{i,1}, \ldots, x^{g}_{i,D}]$ in assumed solution space, where $g$ denotes an iteration number. In each iteration $g$, the population $P^g$ consists of $N^g$ individuals  $[\pmb{x}^{g}_1, \ldots, \pmb{x}^{g}_{N^g}]$ independently subjected to subsequent phases: mutation, crossover, and selection. 

The mutation phase is responsible for random generation of a mutated vector $\pmb{v}^{g}_i$ based on randomly generated scaling factor parameter $F_i^g$ and three randomly chosen individuals ($\pmb{x}^{g}_{pbest_i}$, $\pmb{x}^{g}_{r1_i}$, and $\pmb{x}^{g}_{r2_i}$). The method of determining $F_i^g$ is described in section~\ref{sec:adaptation}. 
Index $pbest_i$ indicates an individual randomly selected from the set of the currently best $N^g\cdot p$ individuals, where $p$ is a parameter.
%Random index 
$r1_i \in \{1, \ldots, N^g\}$ denotes an individual from population $P^g$ and $r2_i \in \{1, \ldots, N^g + |A|\}$, ($r2_i \neq r1_i$) an individual from the union of the population $P^g$ and an external archive $A$. The rules of constructing the external archive $A$ are described in more detail in section~\ref{sec:archive}. 
%The indices $r1_i$ and $r2_i$ must differ from each other ($r1_i \neq r2_i$).
The final form of the mutated vector $\pmb{v}^{g}_i$ can be described as follows:
\begin{equation}\label{eq:mutation}
\pmb{v}^{g}_i = \pmb{x}^{g}_i + F_i^g (\pmb{x}^{g}_{pbest_i} - \pmb{x}^{g}_i) + F_i^g (\pmb{x}^{g}_{r1_i} - \pmb{x}^{g}_{r2_i})
\end{equation}

Next, the parent vector $\pmb{x}^{g}_i$ is crossed with the mutated vector $\pmb{v}^{g}_i$, leading to a trial vector $\pmb{u}^{g}_i = [u^{g}_{i,1}, \ldots, u^{g}_{i,D}]$, in which each element $u_{i,d}^{g}, d=1,\ldots,D$ takes either the value $x_{i,d}^{g}$ (with probability $CR_i^g$) or $v_{i,d}^{g}$, otherwise. 
%The crossover ($u_{i,d}^{g}$ = $v_{i,d}^{g}$) is probabilistic according to the crossover rate $CR_i^g$ parameter. 
The way the $CR_i^g$ is determined is described in section~\ref{sec:adaptation}. Furthermore, one randomly chosen element $u_{i,d_{rand}}^{g}$ is crossed regardless of the probabilistic outcome, i.e.
%Finally, the crossover phase can be represented as follows:
%
\begin{equation}
u_{i,d}^{g}=\begin{cases}
    v_{i,d}^{g} , & \text{if $rand(0,1) \leq CR_i^g$ or $d = d^{rand}_i$}\\
    x_{i,d}^{g} , & \text{otherwise}
\label{eq:crossover}
\end{cases}
\end{equation}
where $rand(0,1)$ denotes a uniformly selected random number from $[0,1)$ and $d^{rand}_i\in \{1, \ldots, D\}$ is 
%an index of the randomly chosen element. 
a randomly selected index. Both $rand(0,1)$ and $d^{rand}_i$ are generated independently for each 
%individual 
$i$.

Finally, the trial vector $\pmb{u}^{g}_i$ is subject to the selection phase. Technically, it is evaluated using fitness function $f$, and its value $f(\pmb{u}_{i}^{g})$ is compared with the value $f(\pmb{x}_{i}^{g})$ of the parent vector $\pmb{x}^{g}_i$. The better vector %remains as a member of 
is promoted to the next generation population 
%for the next generation 
($g+1$). The selection phase can be formally expressed as follows:
\begin{equation}\label{eq:selection}
\pmb{x}_{i}^{g+1}=\begin{cases}
    \pmb{u}_{i}^{g}, & \text{if $f(\pmb{u}_{i}^{g}) < f(\pmb{x}_{i}^{g})$}\\
    \pmb{x}_{i}^{g}, & \text{otherwise}
\end{cases}\end{equation}

LSHADE utilizes a Linear Population Size Reduction (LPSR) mechanism, which results in 
%the dynamic 
changing in time population size $N^g$ during an optimization run. The LPSR mechanism requires $N_{init}$ and $N_{min}$ parameters, indicating the initial and minimal (terminal) population sizes, resp. In each iteration $g$, after the selection phase, the population size $N^{g+1}$ for the next iteration is obtained as follows:
\begin{equation}\label{eq:lpsr}
    N^{g+1} = round \left( \left( \frac{N_{min} - N_{init}}{MAX\_NFE} \right) \cdot NFE + N_{init} \right)
\end{equation}
where $MAX\_NFE$ indicates the optimization budget and $NFE$ is the number of fitness function evaluations made so-far. If the population size is reduced, the worst individuals, in the sense of fitness function value, are removed.

\subsection{External archive}\label{sec:archive}

LSHADE utilizes an external archive $A$ that extends the current population with the individuals (parent vectors) that have been replaced with better offsprings (trial vectors) in the selection phase. 
The archive is of size $|A| = a \cdot N^g$, where $a$ is a parameter. 
%If the archive is full, a randomly selected element is removed to insert a new one in its place.
If the archive is full, a randomly selected element is removed to allow an insertion of a new one in its place.
Likewise, when the archive size is shrunk due to population size reduction, randomly selected elements are removed.

\subsection{Parameter adaptation}\label{sec:adaptation}

The scaling factor $F_i^g$ utilized in the mutation phase (\ref{eq:mutation}), and crossover rate $CR_i^g$ utilized in the crossover phase (\ref{eq:crossover}) are designated using the memory. The purpose of the memory is to store historical values of $F_i^g$  and $CR_i^g$ that succeeded in the selection phase, i.e. the trial vector $\pmb{u}_{i}^{g}$ was better than the parent vector $\pmb{x}_{i}^{g}$.
The memory consists of $H$ pairs of scaling factor entries $M_{F,k}^g$ and crossover rate entries $M_{CR,k}^g$, $k=1,\ldots, H$.
In each iteration, after the selection phase, all successful values of $F_i^g$ and $CR_i^g$ are 
%gathered and 
recorded in sets $S_{F}$ and $S_{CR}$, resp. Then, both sets are transformed using a weighted Lehmer mean so that two values, $mean_{W_L}(S_{F})$ and $mean_{W_L}(S_{CR})$, are obtained.
The following equation describes the weighted Lehmer transformation:

\begin{equation}\label{eq:lehmer}
mean_{W_L}(S) = \frac{\sum^{|S|}_{k=1} w_k S_k^2}{\sum^{|S|}_{k=1} w_k S_k}, \quad\quad w_k = \frac{\Delta f_k}{\sum^{|S|}_{l=1} \Delta f_l}
\end{equation}
where the improvement $\Delta f_p$, $p \in \{k,l\}$ is a difference between the fitness function value of the parent vector ($f(\pmb{x}_{i}^{g})$) and the fitness function value of the trial vector ($f(\pmb{u}_{i}^{g})$).

The memory entries $M_{F,k}^g$ and $M_{{CR},k}^g$ are updated sequentially with values $mean_{W_L}(S_{F})$ and $mean_{W_L}(S_{CR})$, from $k=1$ to $k=H$. After updating the last entry ($k=H$), the updating procedure starts again from $k=1$.

In addition, if all $CR_i^g$ values in set $S_{CR}$ are equal to $0$, the memory updating procedure permanently marks the entry $M_{{CR},k}^g$ with the terminal value $\bot$ (instead of $mean_{W_L}(S_{CR})$).  

% , and remain the entry $M_{{CR},k}^g = \bot$ fixed until the end of the search. The feature is triggered by the situation when all $CR_i^g$ values in set $S_{CR}$ are equal to $0$.}

At the beginning of each iteration, a random index $r_i$ of memory entry is determined, independently for each individual. The values $F_i^g $ and $CR_i^g $ are generated randomly using Cauchy distribution and Normal distribution, resp. with $M_{F,r_i}^g$ and $M_{{CR},r_i}^g$ (taken from the memory) being the parameters of these distributions.
In summary, $F_i^g$ and $CR_i^g$ are generated in the following way:
\begin{equation}\label{eq:cauchy}
F_i^g = rand_{Cauchy}(M_{F,r_i}^g, 0.1)
\end{equation}
\begin{equation}\label{eq:normal}
CR_i^g = \begin{cases} 
0 & \text{if $M_{{CR},r_i}^g = \bot$}\\
rand_{Normal}(M_{CR,r_i}^g, 0.1) & \text{otherwise}
\end{cases}
\end{equation}

If $F_i^g > 1$, $F$ is truncated to $1$, and if $F_i^g \leq 0$, a random generation is repeated.
In case the generated $CR_i^g$ value is outside $[0,1]$, it is truncated to the limit value of $0$ or $1$, resp. 

\section{Proposed pre-screening mechanism}\label{sec:pslshade}

The core of this paper is the proposal of incorporation of the pre-screening mechanism into LSHADE, which results in the psLSHADE method.
The mechanism assumes that each individual $i$ will generate more than one trial vector, but only the most promising one, according to the meta-model evaluation, will be evaluated using the (true) fitness function. 
The proposed pre-screening mechanism includes a modified initial sampling procedure, an archive of samples that stores already evaluated samples, and a global meta-model re-estimated in each iteration. Moreover, all three phases (mutation, crossover and selection) of the underlying LSHADE algorithm have been modified to generate an increased number of trial vectors.

A pseudocode of the psLSHADE algorithm is presented in Algorithm~\ref{alg:pslshade}. A source code 
%of the method and experimental environment 
is available in the public repository
\footnote{\url{https://bitbucket.org/mateuszzaborski/pslshade/}}.

\subsection{Initial population}
The initial population $P^g$ in psLSHADE is generated using Latin Hypercube Sampling~\cite{helton2003latin} to ensure better coverage of the search space compared to purely uniform random sampling. 

\subsection{Genetic operators}
The modified mutation phase generates $N_s$ mutated vectors $\pmb{v}^{g,j}_i$, $j=1,\ldots,N_s$, per individual $i$. All randomly generated values utilized in LSHADE are now designated independently for each of $N_s$ mutated vectors $\pmb{v}^{g,j}_i$. The generation principles, however, remain unchanged, including $M_{F,r_i}^g$ parameter in $F_i^{g,j}$ distribution (\ref{eq:cauchy2}), which is constant in all generation procedures related to individual $i$ in iteration $g$. In summary, the mutated vector $\pmb{v}^{g,j}_i$ is obtained as follows: 
\begin{equation}\label{eq:mutation2}
\pmb{v}^{g,j}_i = \pmb{x}^{g}_i + F_i^{g,j} (\pmb{x}^{g,j}_{pbest_i} - \pmb{x}^{g}_i) + F_i^{g,j} (\pmb{x}^{g,j}_{r1_i} - \pmb{x}^{g,j}_{r2_i})
\end{equation}
where $F_i^{g,j}$ is generated using the Cauchy distribution:
\begin{equation}\label{eq:cauchy2}
F_i^{g,j} = rand_{Cauchy}(M_{F,r_i}^g, 0.1)
\end{equation}

Then, for each individual $i$, $N_s$ trial vectors $\pmb{u}^{g,j}_i = [u^{g,j}_{i,1}, \ldots, u^{g,j}_{i,D}]$, $j=1,\ldots,N_s$, are designated using the same $CR_i^g$ and $d^{rand}_i$ values. 
Also, the value $rand(0,1)$ for individual $i$ is generated once for all $N_s$ trial vectors.
Consequently, each element $u^{g,j}_{i,1}$ is described by the following equation:
\begin{equation}
u_{i,d}^{g,j}=\begin{cases}
    v_{i,d}^{g,j}, & \text{if $rand(0,1) \leq CR_i^g$ or $d = d^{rand}_i$}\\
    x_{i,d}^{g}, & \text{otherwise}
\label{eq:crossover2}
\end{cases}
\end{equation}
where $\forall_{j \in \{1, \ldots, N_s\}}\ CR_i^g = const., d_{rand} = const.$, 
and $rand(0,1) = const.$ 

After the crossover phase, the meta-model's coefficients are estimated. The principles of a global meta-model are described in detail in section \ref{sec:metamodel}. Then, all $N^g \cdot N_s$ trial vectors $\pmb{u}^{g,j}_i$ are evaluated using the meta-model to obtain $N^g \cdot N_s$ surrogate function values $f^{surr}(\pmb{u}^{g,j}_i)$. 
Afterward, the best trial vector $\pmb{u}^{g,best}_i$ is chosen, independently for each individual $i$, based on the $f^{surr}(\pmb{u}^{g,j}_i)$ values.
Finally, $N^g$ best trial vectors are evaluated using the fitness function, and each of them undergoes the selection phase:
\begin{equation}\label{eq:selection2}
\pmb{x}_{i}^{g+1}=\begin{cases}
    \pmb{u}_{i}^{g, best}, & \text{if $f(\pmb{u}_{i}^{g, best}) < f(\pmb{x}_{i}^{g})$}\\
    \pmb{x}_{i}^{g}, & \text{otherwise}
\end{cases}\end{equation}

\subsection{Archive}
The archive of samples stores at most $N_a$ pairs of already evaluated trial vectors $\pmb{u}_{i}^{g,best}$ (or vectors $\pmb{x}_{i}^{0}$ generated by initial sampling) and their respective function values $f(\pmb{u}_{i}^{g,best})$ ($f(\pmb{x}_{i}^{0})$, resp.). A new pair $(\pmb{u}_{i}^{g,best}, f(\pmb{u}_{i}^{g,best}))$ is unconditionally inserted if the archive contains less than $N_a$ elements. If the archive is full, the worst pair 
%$(\pmb{u}_{worst}, f(\pmb{u}_{worst}))$, 
in terms of the fitness function value is removed from the archive, and a new pair is inserted. Removing the worst pair and insertion of a new one is not executed if the worst pair from the archive is better than the new pair.
In order to ensure numerically correct estimation of meta-model parameters, an additional anti-duplicate condition is met. Removing and inserting is not executed if the archive already contains a pair that is similar to a new pair. The pairs are treated as similar when their trial vectors are equal or fitness function values are equal. The accepted numerical tolerance of equality is $10^{-12}$.

\subsection{Parameter adaptation}
The parameter adaptation procedure remains unchanged compared to LSHADE. Utilization of pre-screening is transparent for the memory. All successful pairs ($F_i^{g,best}, CR_i^{g,best}$) that replaced 
%instead of 
($F_i^{g},CR_i^g$) are collected in sets $S_F$ and $S_{CR}$, resp. The remaining steps are analogous to those employed in LSHADE.
%
%{\color{green}Please refer to the supplementary material for our psLSHADE implementation and accompanying experimental environment.}
%
\begin{algorithm}[h]
	\begin{algorithmic}[1]
	\footnotesize
	\State Set all parameters $N_{init}, N_{min}, M_F, M_{CR}, p, a, H, N_a, N_s$ 
	%(Table~\ref{tab:parameters})
	(Section~\ref{sec:params})
    \State Initialize $M_{F,k}^0$ and $M_{CR,k}^0$ memory entries with default values of $M_F$ and $M_{CR}$
    \State Initialize $P^0 = [\pmb{x}_{1}^0, \ldots, \pmb{x}_{N}^0]$ with $N=N_{init}$ 
    %\Comment{
    using Latin Hypercube Sampling
    %} 
    \State Update sequentially the archive of samples with all individuals from $P^0$
    \State $g = 1$
	\While{evaluation budget left}
	\State Generate $N^g \cdot N_s$ mutated vectors $\pmb{v}^{g,j}_i$ using  eq. (\ref{eq:mutation2})
	\State Generate $N^g \cdot N_s$ trial vectors $\pmb{u}^{g,j}_i$ using eq. (\ref{eq:crossover2})
	\State Estimate meta-model parameters (Table~\ref{tab:metamodel})
	\State Calculate $N^g \cdot N_s$ surrogate values $f^{surr}(\pmb{u}^{g,j}_i)$
	\State For each individual $i$ designate the best trial vector $\pmb{u}^{g,best}_i$
	\For{i = 1 to $N^g$}
    	\State Do selection of $\pmb{u}_{i}^{g,best}$ using eq. (\ref{eq:selection2})
    	\State Add $\pmb{u}_{i}^{g,best}$ and $f(\pmb{u}_{i}^{g,best})$ to the archive of samples
	\EndFor
	\State Update memory with $M_{F,k}^g$ and $M_{CR,k}^g$ using eq. (\ref{eq:lehmer})
	\State Set new population size $N^{g+1}$ using eq. (\ref{eq:lpsr})
	\State $g = g + 1$
	\EndWhile
  \caption{Pre-screening LSHADE high-level pseudocode%
  \label{alg:pslshade}}
  \end{algorithmic}
  \end{algorithm}

\subsection{Meta-model characteristic}\label{sec:metamodel}

The meta-model utilized in the pre-screening mechanism is a global linear model that consists of 6 independent transformations of variables: constant, linear, quadratic, modeling interactions, inverse linear, and inverse quadratic. All transformations, including their form and the number of degrees of freedom, are described in Table~\ref{tab:metamodel}.
The final form of the meta-model is a 
%composition 
linear combination of 6~transformations. 
%The resulting meta-model's degrees of freedom $df_{mm}$ equals the sum of degrees of freedom derived from the transformations.
The meta-model coefficients are estimated using Ordinary Least Squares~\cite{weisberg2013applied}.
The meta-model is constructed when the archive of samples contains at least $df_{mm}$ samples. Thus, the archive size should equal at least $df_{mm}$. 
%By default, we assumed $N_a = 2 \cdot df_{mm}$. 

\begin{table}[h]
%\small 
	\caption{A description of transformations and the final meta-model (mm.) The estimated coefficients applied to each variable are omitted for the sake of readability.
    \label{tab:metamodel}}

	\begin{center}%
	    \begin{tabular}{l@{\hskip 0.05in}l@{\hskip 0.05in}l@{\hskip 0.05in}}
        \toprule
             \textbf{Name} & \textbf{Form} & \textbf{DoF}    \\ 
        \midrule
        Constant & $X_{c} = [1]$ & $df_{c} = 1$ \\
        Linear  & $X_{l} = [x_1, \ldots, x_D]$  &  $df_{l} = D$ \\
        Quadratic    & $X_{q} = [x_1^2, \ldots, x_D^2]$  & $df_{q} = D$       \\
        Interactions  & $X_{i} = [x_1x_2, \ldots, x_{D-1}x_D]$ &  $df_{i} = \frac{D(D-1)}{2}$        \\
        Inv. linear  & $X_{il} = [\frac{1}{x_1}, \ldots, \frac{1}{x_D}]$  &  $df_{il} = D$ \\
        Inv. quad.  & $X_{iq} = [\frac{1}{x_1^2}, \ldots, \frac{1}{x_D^2}]$  &  $df_{iq} = D$ \\
        \hline
        Final mm.  & $[X_c + X_l + X_q + X_{i} + X_{il} + X_{iq}]$  &  $df_{mm} = \frac{D^2 + 7D}{2}+1$ \\
          \bottomrule                          
        \end{tabular}
	\end{center}%
\end{table}

\section{Experimental evaluation}\label{sec:experiment}

We evaluated psSHADE on the recent popular 
%The evaluation was made using 
%CEC2021 Special Session on Real-Parameter Single Objective Optimization benchmark suite~\cite{cec2021}, 
\textit{CEC2021 Special Session and Competition on Single Objective Bound Constrained Numerical Optimization benchmark suite}, described in technical report~\cite{cec2021}, henceforth referred to as CEC2021.
CEC2021 consists of 10 bound-constraint functions belonging to the following four categories: unimodal functions ($F_1$), basic functions ($F_2$ - $F_4$), hybrid functions ($F_5$ - $F_7$), and composition functions ($F_8$ - $F_{10}$). Each function is defined for both 10 and 20 dimensions.

In addition, three function transformations are proposed in CEC2021: bias (B), shift (S), and rotation (R). Besides the baseline case without transformations applied, the following four combinations
%, excluding no transformation, 
of transformations (applied to each function) are tested: S, B+S, S+R, and B+S+R.

By default CEC2021 assumes the following optimization budget: $2 \cdot 10^5 \cdot D$ FFEs for 10D problems and $10^6 \cdot D$ FFEs for 20D problems. Please refer to~\cite{cec2021} for 
%more details about CEC2021 and 
explicit function definitions.
%
%%%%%%%%%%%%%%%%%%%%%%%%%%%%%%%%
Since surrogate-assisted algorithms, such as psLSHADE, are generally designed for expensive scenarios, i.e. restricted optimization budgets, we present the results based on the smaller budgets: $10^2 \cdot D$, $10^3 \cdot D$, and $10^4 \cdot D$. 
Employing three different budgets helps to determine when incorporating the pre-screening mechanism is the most effective. 

Each experiment is repeated 30 times. A search range is $[-100,100]^D$ and is the same for all functions, dimensions, and transformations.
In summary, for each optimization budget the entire evaluation process included $10 \times 5 \times 30 \times 2 = 3000$ (functions $\times$ transformations $\times$ repetitions $\times$ dimensions) independent runs.

The performance of the two competing algorithms (baseline LSHADE and MadDE) was examined %using a smaller budget than the benchmark's default.
under the same conditions and in the same manner.
The implementations of LSHADE and MadDE were taken from~\cite{codeoftopmethods}.
%and evaluated them in the same manner as the proposed psLSHADE was evaluated. 
In addition, we made sure that psLSHADE implementation without pre-screening exactly matches the implementation of LSHADE.

\subsection{Performance metrics}
The evaluation metrics are adopted from the aforementioned technical report~\cite{cec2021}.
%{\color{red}For each run, 16 {\color{orange}best-so-far} error values were recorded at different points in time (FFEs).}  
The final score is calculated in the few following steps.
%(i.e. the best one, after the whole budget was exhausted).
%
%The final measure of each algorithm performance as determined in a few steps. 
First, the $SNE$ value is calculated:
\begin{equation}\label{eq:SNE}
SNE = 0.5 \sum_{m=1}^5 \sum_{i=1}^{10} ne_{i,m}^{10D} + 0.5\sum_{m=1}^5 \sum_{i=1}^{10} ne_{i,m}^{20D} 
\end{equation}
where $ne_{i,m}^{dim}$ is a normalized error value (\ref{eq:ne}) for function $F_i$, transformation $m$, and dimension $dim$
%The $ne$ value is calculated independently for each function, transformation and dimension, %in a relative manner 
(for the sake of clarity we omit the indices):
\begin{equation}\label{eq:ne}
ne = \frac{f(\pmb{x}_{best}) - f(\pmb{x}^*)}{f(\pmb{x}_{best})_{max} - f(\pmb{x}^*)}
\end{equation}
where $f(\pmb{x}_{best})$ is the algorithm's best result out of 30 trials (repetitions), $f(\pmb{x}^*)$ is the function's optimal value, and $f(\pmb{x}_{best})_{max}$ is the largest (the worst) $f(\pmb{x}_{best})$ among all algorithms.

The Score1 value is calculated as follows:
\begin{equation}\label{eq:score1}
Score1 = \Biggl(1 - \frac{SNE - SNE_{min}}{SNE}\Biggr) \times 50
\end{equation}
where $SNE_{min}$ is the minimal $SNE$ among all algorithms. Then the $SR$ value is obtained:
\begin{equation}\label{eq:SR}
SR = 0.5 \sum_{m=1}^5 \sum_{i=1}^{10} rank_{i,m}^{10D} + 0.5\sum_{m=1}^5 \sum_{i=1}^{10} rank_{i,m}^{20D} 
\end{equation}
where $rank_{i,m}^{dim}$ is the algorithm's rank among all algorithms for a given function $i$, transformation $m$, and dimension $dim$,
%(10D or 20D), 
using the mean error value from all trials. 

The Score2 value is computed in the following way:
\begin{equation}\label{eq:score2}
Score2 = \Biggl(1 - \frac{SR - SR_{min}}{SR}\Biggr) \times 50
\end{equation}
where $SR_{min}$ is the minimal sum of $rank_{i,m}$ among the compared algorithms.

The final $Score$ measure is the sum of (\ref{eq:score1}) and (\ref{eq:score2}):
%obtained as $Score = Score1 + Score2$.
%
\begin{equation}\label{eq:score}
Score = Score1 + Score2
\end{equation}

For auxiliary measures $SNE$ and $SR$, lower values indicate better performance, whereas for $Score1$, $Score2$, and $Score$, higher values correspond to better results. The maximum value of 
%$Score1$, $Score2$, and 
$Score$ is 100. 
%{\color{red}The best algorithm in terms of $SNE$ and $SR$ at once will gain a $Score = 100$.}

\subsection{Tuning of psLSHADE parameters}\label{sec:params}
%
%Table~\ref{tab:parameters} presents psLSHADE parameters.
%
\subsubsection{Shared psLSHADE/LSHADE parameters.}

For the sake of fair comparison, all psLSHADE parameters shared with LSHADE have the same values in both algorithms, and follow the parameterization used previously for benchmarking LSHADE on CEC2021~\cite{codeoftopmethods}.
%To ensure the quality of comparisons, psLSHADE was parameterized analogous to LSHADE. 
%
The following parameters are shared by both algorithms:
initial population size $N_{init} = 18 \cdot D$, final population size $N_{min} = 4$, initial value of $M_F = 0.5$, initial value of $M_{CR} = 0.5$,
best rate $p = 0.11$, archive rate $a = 1.4$, memory size $H = 5$.

\subsubsection{psLSHADE pre-screening component parameters}
%The following two parameters are specific for the psLSHADE:
%archive size $N_a = 2 \cdot df_{mm}$, trial vectors per individual $N_s = %\{2,5,10,20\}$.
%
%\begin{table}[h]
%	\caption{LSHADE and pre-screening component parameters.%
%	\label{tab:parameters}}
%	\begin{center}
%	\begin{tabular}{lr}
%		\multicolumn{2}{l}{\textbf{LSHADE parameters}} \\
%		Initial population size $N_{init}$ & $18 \cdot D$ \\
%		Final population size $N_{min}$ & $4$ \\
%		Initial $M_F$ = Initial $M_{CR}$ & $0.5$ \\
%	    Initial $M_{CR}$ & $0.5$ \\
%	    Best rate $p$ & $0.11$ \\
%	    Archive rate $a$ & $1.4$ \\
%	    Memory size $H$ & $5$ \\
%		\multicolumn{2}{l}{\textbf{Pre-screening component parameters}} \\
%		Archive size $N_a$ & $2 \cdot df_{mm}$ \\
%		Trial vectors per individual $N_s$ & $\{2,5,10,20\}$ \\
%	\end{tabular}%
%	\vspace{-1em}%
%\end{center}%
%\end{table}
%

The two additional psLSHADE parameters are $N_a$ and $N_s$. The archive size ($N_a$) was selected arbitrary (with no tuning) following the lq-CMA-ES~\cite{hansen2019global} parameterization. In~\cite{hansen2019global} the archive is of the size $max(\lambda, 2 \cdot df_{max})$, where $\lambda$ is the population size and $df_{max}$ is the number of degrees of freedom of the most complex meta-model. 
Due to population size reduction in psLSHADE, we conditioned the archive size on the number of degrees of freedom only.

For the number of trial vectors per individual ($N_s$) the tuning tests were performed with $10^3 \cdot D$ budget and the following values: $N_s = \{2,5,10,20\}$.
%
%Selected configurations were compared with each other using the benchmark procedure. The purpose of the experiment was to select the best configuration to be compared afterward to the baseline LSHADE and MadDE. 
%Besides, the hypothesis that there is an $N_s$ value above which the algorithm's performance starts to decrease needs to be verified.
%
psLSHADE results obtained for all four $N_s$ choices are presented in Table~\ref{tab:resultsurr}. The best performing value was $N_s = 5$. 
%The configuration with 
$N_s=2$ was a slightly weaker choice, indicating that increasing the number of trial vectors up to a certain point ($N_s=5$ in this case) improves performance, rendering the use of pre-screening beneficial.
Both $N_s=10$ and $N_s=20$ appeared to be inferior selections. 
%have proven to be less efficient than 
%the one with 
%$N_s=5$. 
Moreover, $N_s=20$ performed much worse than $N_s=10$, which clearly outlined a trend of decreasing performance beyond a certain threshold.
%while $N_s$ increases overly. 
%
%The hypothesis assuming the existence of such a performance threshold has been confirmed. 

At the same time we would like to point out that the above tuning procedure was by no means exhaustive and we would rather advocate for a certain qualitative performance pattern with respect to $N_s$, than for particular $N_s$ values. In conclusion, $N_s=5$ was considered sufficient and used in further comparisons. 
%has not been designated strictly. However, it was not the purpose of the comparison. The parameter $N_s=5$ was found to be the best and will be used in further comparisons.
%
%Nevertheless, the noted trend has inspired in-depth research of the meta-model's behavior %(see Section \ref{sec:convergence} and Section \ref{sec:analysis}). 

\begin{table}[h]
	\caption{Scores achieved by psLSHADE with $N_s = \{2,5,10,20\}$ for $10^3\cdot D$ optimization budget.
    \label{tab:resultsurr}}
	\begin{center}%
	    \begin{tabular}{r|r|r|r|r}
        \toprule
             \textbf{\diagbox{Score}{Algo}} & \textbf{$N_s = 2$} & \textbf{$N_s = 5$} & \textbf{$N_s = 10$} & \textbf{$N_s = 20$}     \\ 
        \midrule
        SNE & 34.87 & 30.80 & 33.59 & 41.42 \\
        SR & 101.25 & 93.75 & 133.75 & 171.25\\
        Score 1 & 44.17 & 50.00 & 45.84 & 37.18 \\
        Score 2 & 46.30 & 50.00 & 35.05 & 27.37 \\
        \textbf{Score} & \textbf{90.46} & \textbf{100.00} & \textbf{80.89} & \textbf{64.56} \\
          \bottomrule                          
        \end{tabular}
	\end{center}%
\end{table}

\subsection{Comparison with LSHADE and MadDE}\label{tab:comparison}

A comparison of psLSHADE with LSHADE and MadDE was carried out with three optimization budgets: $10^2 \cdot D$, $10^3 \cdot D$, and $10^4 \cdot D$. MadDE, utilizing the same CEC2021 benchmark but with a default optimization budget, outperformed several other algorithms (AGSK~\cite{mohamed2020evaluating}, LSHADE~\cite{tanabe2014improving}, LSHADE\_cnEpSin~\cite{awad2017ensemble}, j2020~\cite{brest2020differential}, and IMODE~\cite{sallam2020improved}) in a comparison reported by the MadDE authors~\cite{biswas2021improving}.
Hence, due to its superior performance, MadDE presents itself as a strong 
%and demanding 
competitive method.
%offers a suitable benchmark for the proposed psLSHADE.
A comparison with the baseline LSHADE, also being a highly efficient approach, directly verifies the value of the proposed pre-screening enhancement.
%is, excluding other factors.

The comparison results for all three optimization budgets 
%($10^2 \cdot D$, $10^3 \cdot D$, and $10^4 \cdot D$) 
are shown in Table~\ref{tab:results102}. 
%The proposed 
psLSHADE distinctly outperforms both LSHADE and MadDE with the smallest budget of $10^2 \cdot D$. The differences are significant, especially when looking at the partial measures $SNE$ and $SRE$. Moreover, being superior in non-expensive optimization budgets, MadDE is in this case less efficient than baseline LSHADE. 
A highly restricted optimization budget might induce the mediocre performance of MadDE due to instability of some of its adaptation procedures.
% characterizing MadDE.

In the regime of $10^3 \cdot D$ budget the results are qualitatively similar to those obtained for $10^2 \cdot D$ budget. psLSHADE still outperforms both competitors, although its advantage is not so prevailing. MadDE remains the least efficient approach.

The $10^4 \cdot D$ optimization budget can be classified as a borderline between expensive and non-expensive scenarios. 
%Nevertheless, we conducted an experimental evaluation also in this case to catch the performance trends for the examined algorithms with further increasing optimization budget. 
In this case, all algorithms perform at a similar level, with MadDE being slightly less effective than the remaining two methods. Considering the \textit{Score} metrics, psLSHADE seems to be marginally less efficient than LSHADE, however, its $SNE$ value outperforms that of LSHADE. 
A relatively weaker psLSHADE 
%optimization 
performance for larger 
%optimization 
budgets may be caused by the decreasing model fit, as the number of evaluations increases. This phenomenon is illustrated further in Section~~\ref{sec:analysis}.

\begin{table}[ht]
	\caption{Scores achieved by MadDE, LSHADE and psLSHADE with $N_s = 5$ and $10^2\cdot D$, $10^3\cdot D$, $10^4\cdot D$ optimization budgets.
    \label{tab:results102}}
	\begin{center}%
	    \begin{tabular}{r|r|r|r|r}
        \toprule
             \textbf{\diagbox{O. b.}{Algo}} & \textbf{Score} & \textbf{MadDE} & \textbf{LSHADE} & \textbf{psLSHADE}    \\ 
        \midrule
        \multirow{5}{*}{$10^2 \cdot D$} & SNE & 41.72 & 32.38 & 19.58 \\
        & SR & 131.00 & 110.50 & 58.50  \\
        & Score 1 & 23.46 & 30.23 & 50.00  \\
        & Score 2 & 22.33 & 26.47 & 50.00 \\
        & \textbf{Score} & \textbf{45.79} & \textbf{56.70} & \textbf{100.00}  \\
        \hline
        \multirow{5}{*}{$10^3 \cdot D$} & SNE & 37.92 & 25.38 & 20.36 \\
        & SR & 125.75 & 104.50 & 69.75 \\
        & Score 1 & 26.85 & 40.10 & 50.00  \\
        & Score 2 & 27.73 & 33.37 & 50.00 \\
        & \textbf{Score} & \textbf{54.58} & \textbf{73.48} & \textbf{100.00}  \\
        \hline
        \multirow{5}{*}{$10^4 \cdot D$} & SNE & 28.50 & 26.81 & 26.37 \\
        & SR & 105.50 & 94.00 & 100.50  \\
        & Score 1 & 46.26 & 49.18& 50.00  \\
        & Score 2 & 44.55 & 50.00 & 46.77 \\
        & \textbf{Score} & \textbf{90.81} & \textbf{99.18} & \textbf{96.77}  \\
          \bottomrule                          
        \end{tabular}
	\end{center}%
\end{table}

To sum up, psLSHADE is highly effective in scenarios with restricted FFE budgets ($10^2 \cdot D$ and $10^3 \cdot D$), leaving both competitors: LSHADE and MadDE behind. In the case of $10^4 \cdot D$ budget, psLSHADE is narrowly worse than LSHADE and slightly more effective than MadDE.

\subsection{Computational complexity}
All experiments were conducted using the following system setup. OS: Windows 10, CPU: Intel Core i7-4700MQ (2.40Ghz), RAM: 16GB, Language: Matlab R2020a, Compiler: MinGW64 C/C++ Compiler.
%\begin{itemize}
%    \item OS: Windows 10
%    \item CPU: Intel Core i7-4700MQ (2.40Ghz)
%    \item RAM: 16GB
%    \item Language: Matlab R2020a
%    \item Compiler: MinGW64 C/C++ Compiler
%\end{itemize}

psLSHADE and LSHADE complexity was computed according to the CEC2021 technical report~\cite{cec2021} and 
%The experimental complexity of both algorithms 
is presented in Table~\ref{tab:complexity}. 
%The proposed 
Computational complexity of psLSHADE is clearly higher than that of LSHADE (by a factor of $25.8$ and $72.8$ for 10D and 20D functions, resp.) which is caused by additional operations stemming from pre-screening utilization.

At the same time, it should be underlined that psLSHADE complexity does not increase with the number of FFEs.
Since the meta-model utilizes the archive of samples of limited size of $N_a$, after the first $N_a$ FFEs, all algebraic operations related to meta-model estimation maintain a constant complexity.
Therefore, the requested optimization time, counted in seconds, is linearly dependent on the FFE budget.

%{\color{orange}\st{The authors indicate there is room for improving the computational complexity of psLSHADE by optimizing some algebraic operations in the implementation.}}
On a general note, the \textit{``overhead''} of psLSHADE optimization time (excluding the time of FFEs) with $2 \cdot 10^5$ FFEs, equals approx. $34s / 83s$ for a $10D / 20D$ function, resp. ($T_2-T_1$ in Table~\ref{tab:complexity}). Considering the perspective of expensive optimization (where FFEs are typically costly), the reported times seem acceptable.

\begin{table}[h]
	\caption{Computational complexity of psLSHADE and LSHADE calculated according to~\cite{cec2021} (i.e. for one benchmark representative). $T_0$ is the time of a test program run. $T_1$ - time of pure $2 \cdot 10^5$ FFEs of $F_1$ function. $T_2$ - the average running time of the algorithm for $F_1$ with $2 \cdot 10^5$ evaluation budget. $(T_2 - T_1)/ T_0$ is the final complexity. 
	%Please consult~\cite{cec2021} for further explanation.
    \label{tab:complexity}}
	\begin{center}%
	    \begin{tabular}{r|r|r|r|r|r}
        \toprule
            \textbf{$D$} &  \textbf{$Algorithm$} & \textbf{$T_0 [s]$} & \textbf{$T_1 [s]$} & \textbf{$T_2 [s]$} & \textbf{$(T_2 - T_1)/ T_0$}     \\ 
        \midrule
        \multirow{2}{*}{10} & psLSHADE & \multirow{4}{*}{0.004841} & 11.3341  & 44.8177 & 6916.6669\\
                            & LSHADE &                             & 0.2027   & 1.5016 & 268.3196\\
        \multirow{2}{*}{20} & psLSHADE &                           & 10.9902  & 93.9839 & 17143.9248 \\
                            & LSHADE &                             & 0.2040   & 1.3442 & 235.5063\\

          \bottomrule                          
        \end{tabular}
	\end{center}%
\end{table}

\subsection{Convergence of psLSHADE}\label{sec:convergence}

The experimental results indicate that the pre-screening mechanism is most advantageous in the case of $10^2 \cdot D$ optimization budget. 
%Therefore, the performance decrease is expected to be observable during the $10^3 \cdot D$ optimization budget.
However, for the sake of generality of the presented observations, in further analysis of psLSHADE performed in the reminder of the paper a bigger (i.e. less favourable) budget of $10^3 \cdot D$ is considered.

\begin{figure*}[h!]
	\begin{center}
		\includegraphics[width=1.0\textwidth]%
		{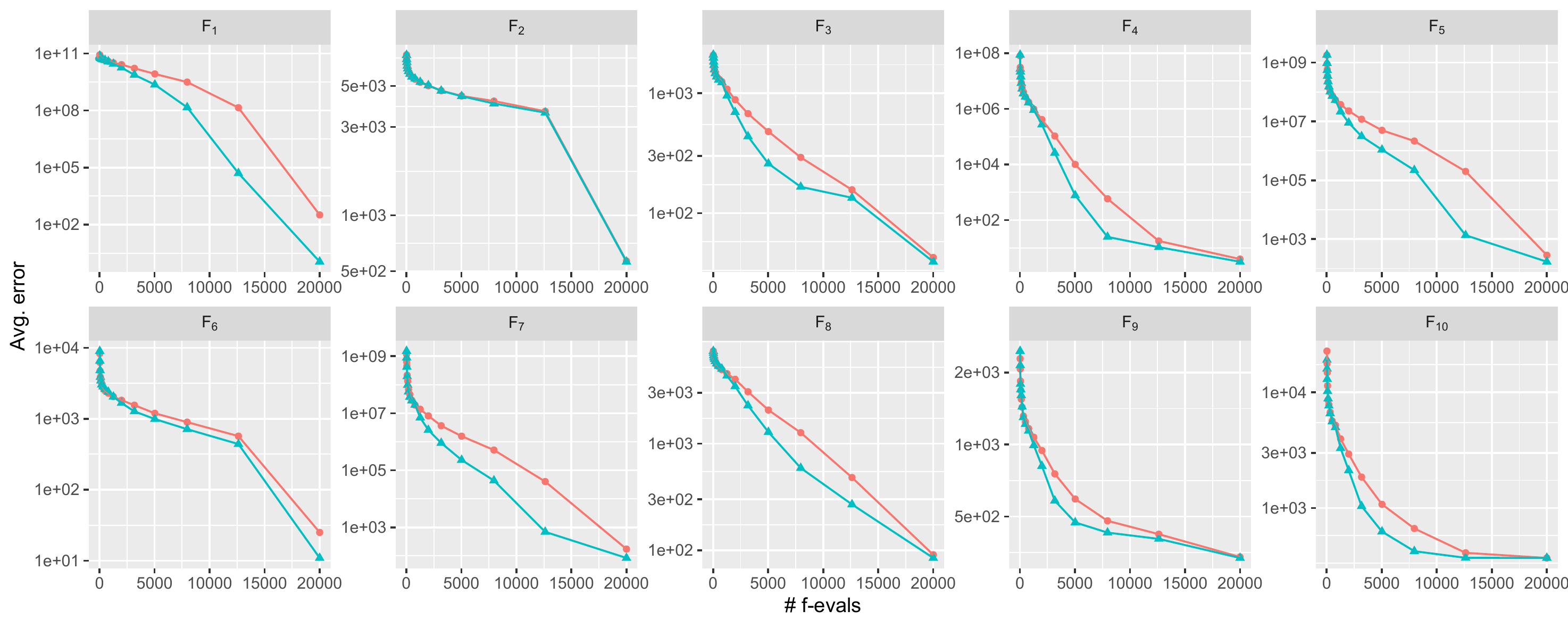}
	\end{center}
    \vspace*{-5mm}
	\caption{The averaged convergence of psLSHADE (blue line) and LSHADE (red line) by function in 20D with $10^3 \cdot D$ optimization budget. The x-axis represents the number of FFEs, and the y-axis the average error (a difference from the optimum). 
	\label{fig:convergence20}}
\end{figure*}

Following~\cite{cec2021} we recorded 16 error values ($f(\pmb{x}) - f(\pmb{x}^*)$) for each optimization run after certain FFE counts. For each function, we obtained 150 vectors (5 transformations $\times$ 30 repetitions) of error values. Figure~\ref{fig:convergence20} presents the convergence of psLSHADE and LSHADE, 
%The convergence is 
calculated as the average error value after each FFE count point, 
%and is presented 
separately for each 20D function. The respective plots for 10D functions are presented in the supplementary material.

%In addition, we performed a 
%According to Mann–Whitney 
%statistical 
%test 
%investigating if the final solution found is significantly better for psLSHADE compared to LSHADE. For 
In $77$ of $100$ test cases (functions $\times$ transformations $\times$ dimensions), the difference between psLSHADE and LSHADE was significant 
%at $0.05$ \textit{p-value} threshold
(Mann–Whitney test, \textit{p-value}=$0.05$). 
%There was no case in which 
In none of the cases psLSHADE was significantly worse than LSHADE.
The results demonstrate that superior convergence of psLSHADE over LSHADE can be observed for $9$ out of $10$ functions. 
It is worth noting that an improved convergence is especially noticeable in the initial optimization phase, depending on the function until $5000$-$15000$ FFEs. In the final optimization phase, the difference starts to diminish, on average.

\begin{figure}[ht]%
    \centering
    \subfloat[\centering Bent Cigar Function]{{\includegraphics[width=3.8cm]{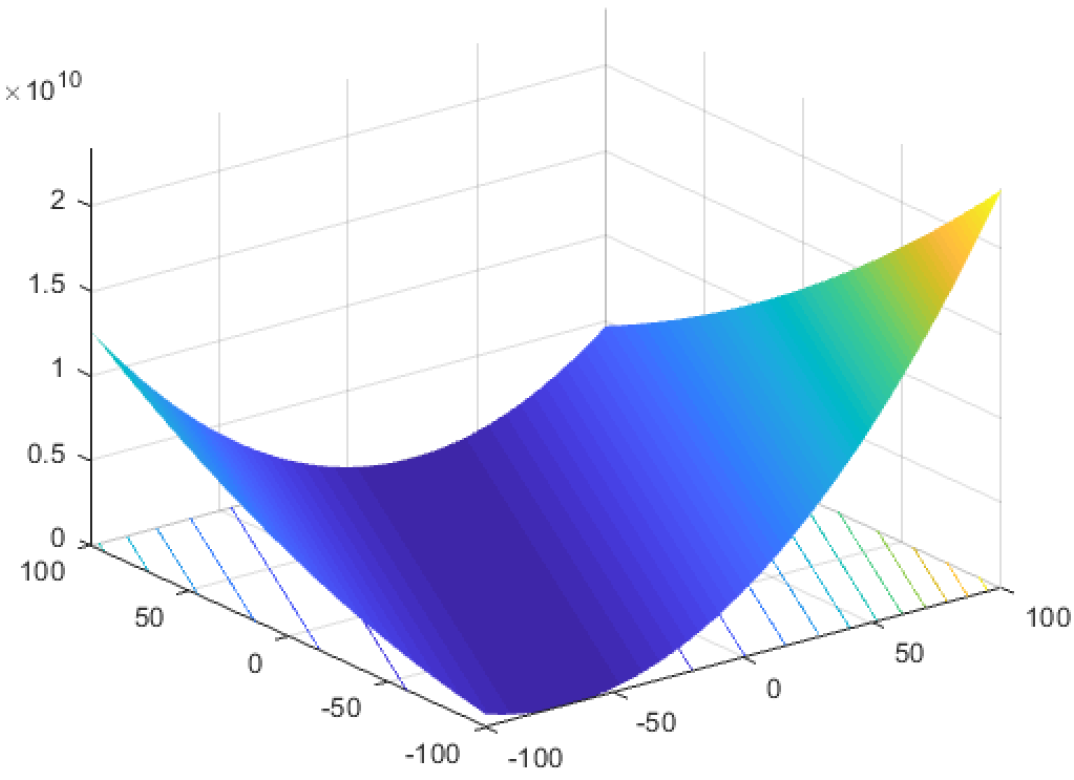} }}%
    \qquad
    \subfloat[\centering Shifted and Rotated Schwefel’s Function]{{\includegraphics[width=3.8cm]{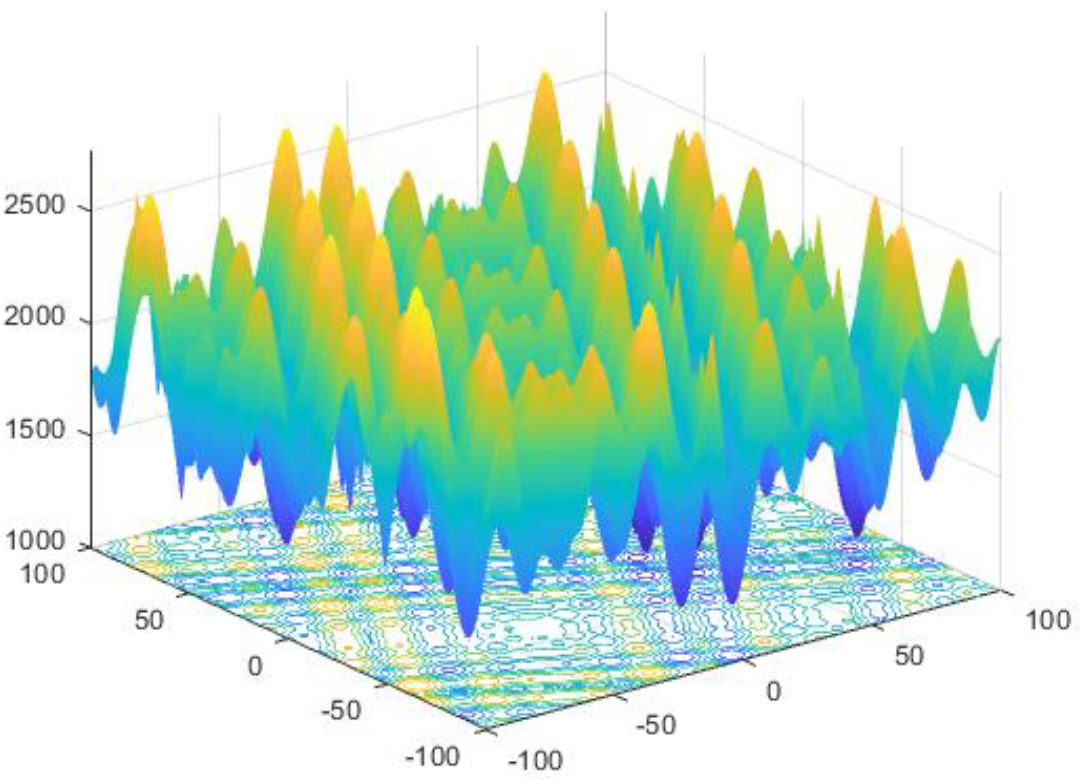} }}%
    \caption{The 3D maps of $2D$ versions of functions $F_1$ (a) and $F_2$ (b). The figures are reprinted from~\cite{cec2021}.}%
    \label{fig:f2f2}%
\end{figure}

The most notable increase of convergence occurs for function $F_1$, whereas for function $F_2$ the advantage is negligible. 
The 2D versions of both functions are presented in Figure~\ref{fig:f2f2}. $F_1$ is unimodal and smooth, while $F_2$ is multi-modal with a huge number of local optima. Moreover, $F_2$ does not possess a visible global structure, making the global meta-model not so much useful in this case. 
Nevertheless, the meta-model does not lead to premature convergence, which is always a risk with such ill-conditioned functions.

The convergence based on error values is a key performance indicator, however, it does not explain how the pre-screening mechanism affects the population. Hence, we designed an experiment in which we measured the hyper-volume (h-v) of the population in each generation $g$. The h-v represents the volume of a hyperrectangle spanning a population and
helps to estimate the population dispersion.
%or converged to some optimum, in general local. 
Formally, the h-v is calculated as follows:
\begin{equation}\label{eq:hypervolume}
h^g = \prod_d^{D} \Bigl(max(x^{g}_{i,d}) - min(x^{g}_{i,d})\Bigr)
\end{equation}

\begin{figure}[ht]
	\begin{center}
		\includegraphics[width=0.48\textwidth]%
		{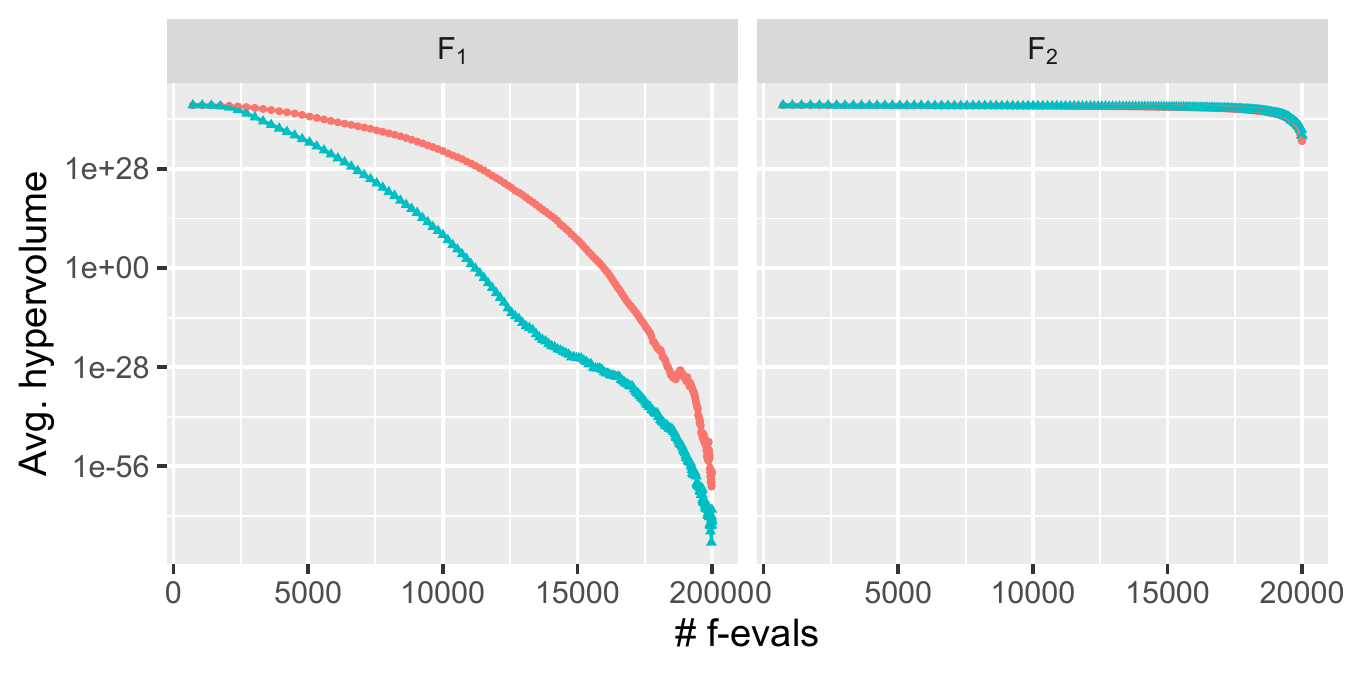}
	\end{center}
    \vspace*{-5mm}
	\caption{The hyper-volume (h-v) of psLSHADE (blue line) and LSHADE (red line) for $F_1$ and $F_2$ in 20D with $10^3 \cdot D$ optimization budget. The x-axis represents the number of FFEs, and the y-axis the average h-v in each generation. 
	\label{fig:hypervolumef1f2}}
\end{figure}

Figure~\ref{fig:hypervolumef1f2} demonstrates changes of the h-v sizes of psLSHADE and LSHADE during an optimization run for $F_1$ and $F_2$. Measurements were collected in each generation, but for consistency, the figure refers to the number of FFEs made so far in each generation. For each function, in a given generation, the h-v value represents the average of 150 values (5 transformations $\times$ 30 repetitions).
%per function and generation from 150 values (5 transformations $\times$ 30 repetitions).
%
%
%The presented comparison shows 
The figure
%~\ref{fig:hypervolumef1f2} 
confirms that significant convergence increase in psLSHADE vs.
%compared to 
LSHADE (presented in Figure \ref{fig:convergence20}) is strongly correlated with the h-v decrease.
%The observation confirms that 
For function $F_1$, with a substantial global structure, the use of pre-screening results in generally faster convergence of each individual (to the global optimum), which results in the h-v reduction.
No significant change of h-v is observed for $F_2$, which confirms the assumption that in this case the pre-screening did not cause premature convergence to a local optimum, i.e. the search ability of psLSHADE has been preserved.
%the ability of LSHADE's searchability. 
The respective figures for the remaining functions in both 10D and 20D versions are included in the supplementary material.

\section{Meta-model performance}\label{sec:analysis}

Consistent with the experimental evaluation, the usefulness of incorporating the pre-screening mechanism into LSHADE has been demonstrated. Notwithstanding, the convergence plots signify some issues with the meta-model in the final optimization phase. For most functions, the advantage accumulated earlier begins to diminish at the end of $10^3 \cdot D$ optimization budget. Two hypotheses regarding this phenomenon emerge. The first one assumes that the global-meta model is well-fitted at the beginning due to the regular dispersion of the initial population. However, as the population loses regularity during the optimization run, the meta-model becomes not fitted correctly, i.e. its selection of trial vectors behaves more randomly. Please recall that the meta-model utilizes the archive containing $N_a$ best-so-far evaluated samples. Consequently, the global meta-model may tend towards the local model. 

The other conjecture is a more pessimistic version of the first one and assumes that the meta-model not only tends
%starts 
to behave randomly, but at some point may even start to select worse-than-average trial vectors.

In order to verify these conjectures, we examined the accuracy of the meta-model during an optimization run. In a dedicated experiment we recorded whether the trial vector $\pmb{u}_{i}^{g, best}$ selected from all $N_s$ trial vectors $\pmb{u}_{i}^{g, j}$ for FFE was, indeed, the best one in terms of the fitness function value (not only the meta-model estimation). 
%Although the fitness function was used as a ranking measure, {\color{blue}it did not disturb} the optimization run and final results.
Each iteration generated $N^g$ logical values: \textit{true} or \textit{false}, referring to $N^g$ individuals. 
%%%%%%%%%%%%%
For each function and each iteration, there were $150 \cdot N^g$ (5 transformations $\times$ 30 repetitions $\times N^g$) logical values obtained. Calculating the share of \textit{true} values in all gathered values leads to the average meta-model accuracy per function and iteration.
Figuratively speaking, the perfect meta-model will gain the average accuracy of $1$ in each iteration. 
In contrast, an entirely random meta-model will achieve the average accuracy $\frac{1}{N^s}$.

\begin{figure}[h]
	\begin{center}
		\includegraphics[width=0.48\textwidth]%
		{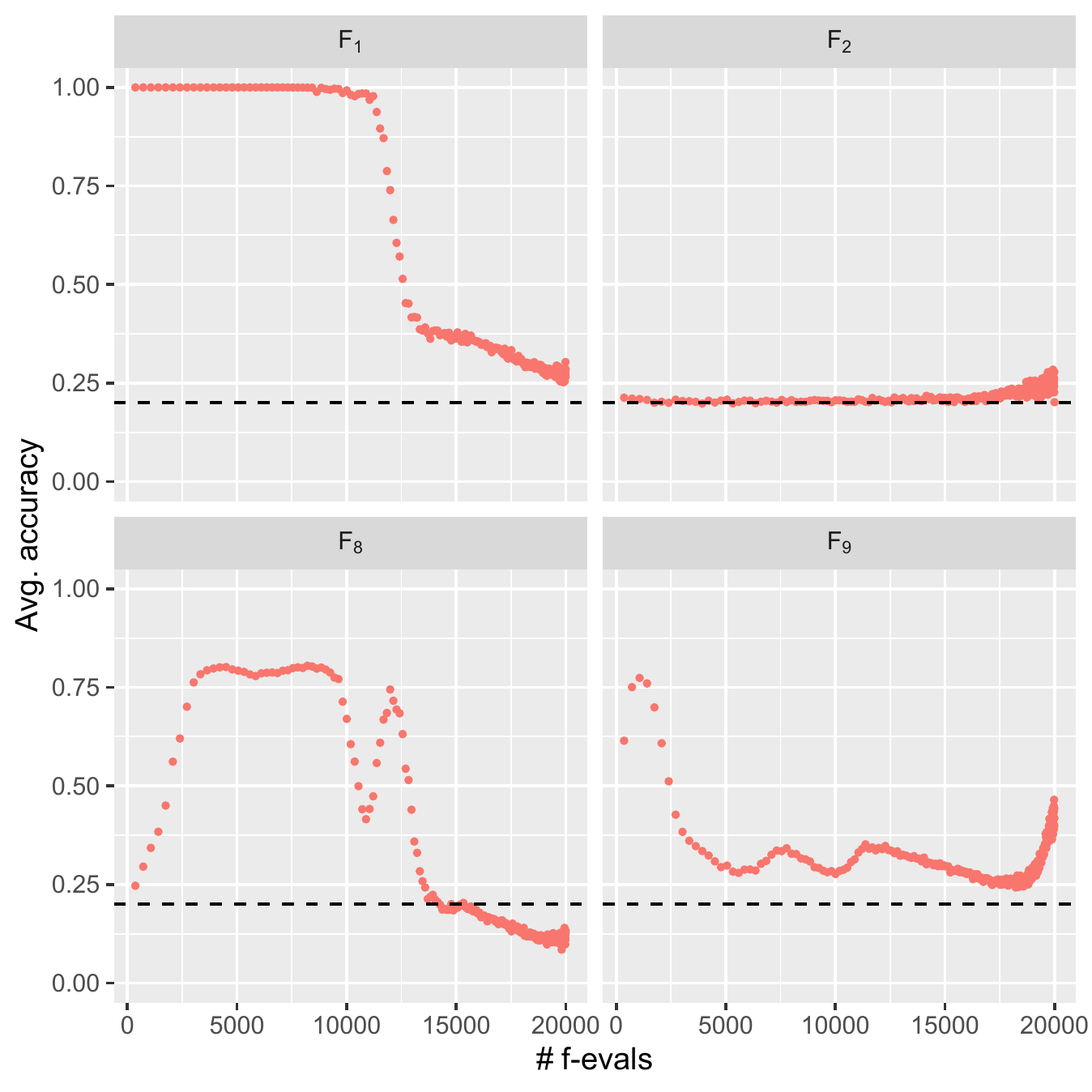}
	\end{center}
    \vspace*{-5mm}
	\caption{The average accuracy of meta-model selections in each generation for functions $F_1$, $F_2$, $F_8$ and $F_9$ in 20D with $10^3 \cdot D$ optimization budget. The x-axis represents the number of FFEs and the y-axis the average accuracy.
	\label{fig:accuracyf1f2f8f9}}
\end{figure}

Figure~\ref{fig:accuracyf1f2f8f9} presents the average meta-model accuracy for functions $F_1$, $F_2$, $F_8$ and $F_9$ in 20D. 
For $F_1$, the accuracy is close to 1 in most iterations and never falls below the randomness threshold ($0.2$). However, a significant drop 
%from the accuracy 
of nearly $0.7$ after $1.2 \cdot 10^4$ evaluations can be observed. This decline may be caused by the convergence of some trials to global optimum earlier than after $10^3 \cdot D$, in which case the meta-model has become irrelevant.

For $F_2$, the meta-model accuracy is as expected. The proposed global meta-model was not able to estimate the ill-conditioned function properly and pre-screen samples better than randomly. Towards the end, the average accuracy began to increase. The reason for this phenomenon may be convergence to some local optimum.

$F_8$ and $F_9$ were chosen 
%for comparison 
because their convergence plots indicated a notable decrease in meta-model accuracy in the final phase of the optimization run.
The meta-model accuracy for $F_8$ remains significantly above the randomness threshold until approx. $1.4 \cdot 10^4$ FEEs. Contrasting this observation with the convergence plot (Figure~\ref{fig:convergence20}), we observe a similar point ($\approx 1.4 \cdot 10^4$) when the convergence, compared to LSHADE, starts to decline. Regardless, the accuracy of the model after $1.5 \cdot 10^4$ FFEs is concerning. It decreased below the randomness threshold. At the end of the budget, the convergence plots of psLSHADE and LSHADE almost meet. 

For $F_9$, a significant reduction in meta-model accuracy occurred earlier, so the convergence advantage of psLSHADE over LSHADE also started to decline earlier.

%{\color{red}\st{The observations presented above suggest a strong relationship between model profitability and its performance.}}
%

The real-time evaluation of the meta-model performance could be beneficial for two reasons. Firstly, deactivation of useless, in particular cases, pre-screening can reduce the computational complexity (see Table~\ref{tab:complexity}). Secondly, the $F_8$ case showed that meta-model predictions might sometimes be worse than random, so deactivation of the meta-model seems even more reasonable in this case.

Unfortunately, the \emph{true/false} accuracy measure presented above 
%in this section 
is not suitable for real-world implementations due to the $5$ times higher number of costly FFEs.
Therefore, we conducted a related experiment but used other performance metrics.
%benchmarks.
The first criterion was the well-known coefficient of determination $R^2 \in [0,1]$. In each iteration, its value was designed after the meta-model had been fitted. The second criterion was Kendall's $\tau \in [0,1]$, measuring the rank correlation between $N^g$ fitness function values $f(\pmb{u}_{i}^{g, best})$ and their respective meta-models estimates $f^{surr}(\pmb{u}_{i}^{g, best})$. Kendall's $\tau$, like $R^2$, was determined in each generation. Importantly, both measures are cost-free in terms of FFEs (no auxiliary evaluations are required).

\begin{figure}[h]
	\begin{center}
		\includegraphics[width=0.48\textwidth]%
		{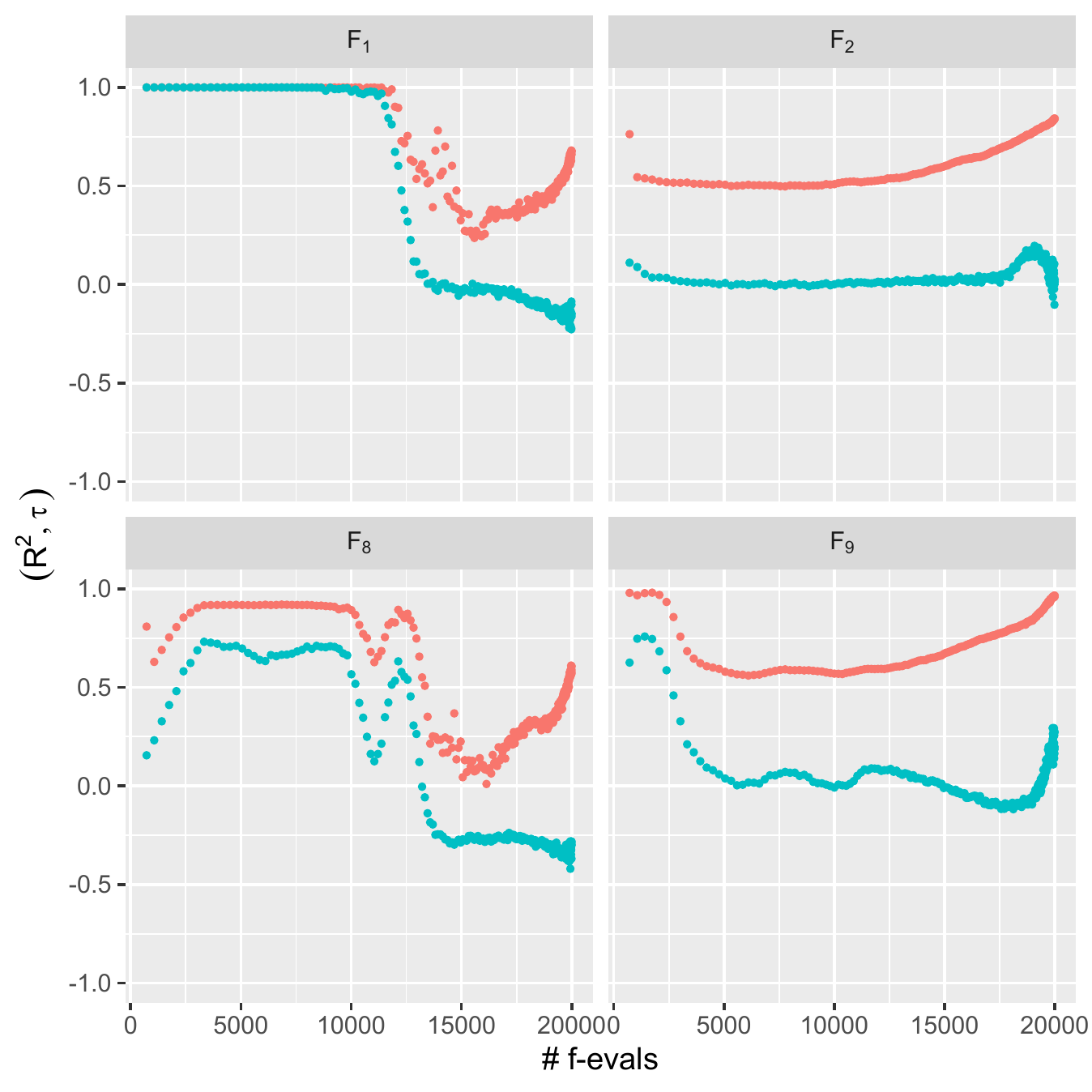}
	\end{center}
    \vspace*{-5mm}
	\caption{The average $R^2$ and Kendall's $\tau$ in each generation for functions $F_1$, $F_2$, $F_8$ and $F_9$ in 20D with $10^3 \cdot D$ optimization budget. The x-axis represents the number of FFEs and the y-axis the average values of $R^2$ and Kendall's $\tau$.
	\label{fig:r2tauf1f2f8f9}}
\end{figure}

The results obtained for the same set of $4$ functions are presented in Figure~\ref{fig:r2tauf1f2f8f9}.
The main disadvantage of $R^2$ is its uncertain impact on the final performance, i.e. the meta-model can be overfitted or inaccurate because of a small number of samples utilized in coefficient estimation. 
%Kendalls's $\tau$, on the other hand, is a measure designated in the previous iteration when, in general, the population was different. 
%
Nonetheless, for both criteria ($R^2$ and Kendall's~$\tau$) the results are highly similar in shape to the accuracy measure.
Therefore, we assume they can potentially be helpful in conditional disabling of the meta-model in practical applications.

\section{Conclusion}\label{sec:conclusions}
In this paper we introduced the psLSHADE algorithm that enhances the well-known LSHADE method with the pre-screening mechanism. On a popular CEC2021 benchmark, the proposed method outperformed LSHADE and MadDE in expensive scenarios (restrictive budget of $10^2 \cdot D$ and $10^3 \cdot D$ FFEs). In $10^4 \cdot D$ budget, psLSHADE performed on par with LSHADE and continued to outperform MadDE.
%The evaluation was made using the CEC2021 benchmark. 

Additionally, a comprehensive analysis of the meta-model performance is presented,
%in the paper, 
including accuracy, fit ($R^2$), and rank correlation (Kendall's $\tau$). 
We also investigate in which cases the pre-screening is indeed beneficial and how does it affect the population in terms of its h-v.
Finally, we demonstrate that the meta-model performance can potentially be monitored in real-time, so as to activate the pre-screening solely when advantageous.

The future work concerns further analysis and development of deactivation conditions of the meta-model (extending the remarks presented at the end of section~\ref{sec:analysis}). 
%as well as tuning or online adaptation of the pre-screening parameters in the context of the optimization budget size.

\section*{Acknowledgments}
Studies were funded by BIOTECHMED-1 project granted by Warsaw University of Technology under the program Excellence Initiative: Research University (ID-UB).

\bibliographystyle{ACM-Reference-Format.bst}
\bibliography{bib}


\begin{thebibliography}{32}

%%% the Web address.

\ifx \showCODEN    \undefined \def \showCODEN     #1{\unskip}     \fi
\ifx \showDOI      \undefined \def \showDOI       #1{#1}\fi
\ifx \showISBNx    \undefined \def \showISBNx     #1{\unskip}     \fi
\ifx \showISBNxiii \undefined \def \showISBNxiii  #1{\unskip}     \fi
\ifx \showISSN     \undefined \def \showISSN      #1{\unskip}     \fi
\ifx \showLCCN     \undefined \def \showLCCN      #1{\unskip}     \fi
\ifx \shownote     \undefined \def \shownote      #1{#1}          \fi
\ifx \showarticletitle \undefined \def \showarticletitle #1{#1}   \fi
\ifx \showURL      \undefined \def \showURL       {\relax}        \fi
% invisible to TeX
\providecommand\bibfield[2]{#2}
\providecommand\bibinfo[2]{#2}
\providecommand\natexlab[1]{#1}
\providecommand\showeprint[2][]{arXiv:#2}

\bibitem[\protect\citeauthoryear{Auger and Hansen}{Auger and Hansen}{2005}]%
        {auger2005restart}
\bibfield{author}{\bibinfo{person}{Anne Auger} {and} \bibinfo{person}{Nikolaus
  Hansen}.} \bibinfo{year}{2005}\natexlab{}.
\newblock \showarticletitle{A restart CMA evolution strategy with increasing
  population size}. In \bibinfo{booktitle}{\emph{2005 IEEE congress on
  evolutionary computation}}, Vol.~\bibinfo{volume}{2}. IEEE,
  \bibinfo{pages}{1769--1776}.
\newblock


\bibitem[\protect\citeauthoryear{Auger, Schoenauer, and Vanhaecke}{Auger
  et~al\mbox{.}}{2004}]%
        {auger2004ls}
\bibfield{author}{\bibinfo{person}{Anne Auger}, \bibinfo{person}{Marc
  Schoenauer}, {and} \bibinfo{person}{Nicolas Vanhaecke}.}
  \bibinfo{year}{2004}\natexlab{}.
\newblock \showarticletitle{LS-CMA-ES: A second-order algorithm for covariance
  matrix adaptation}. In \bibinfo{booktitle}{\emph{International Conference on
  Parallel Problem Solving from Nature}}. Springer, \bibinfo{pages}{182--191}.
\newblock


\bibitem[\protect\citeauthoryear{Awad, Ali, and Suganthan}{Awad
  et~al\mbox{.}}{2017}]%
        {awad2017ensemble}
\bibfield{author}{\bibinfo{person}{Noor~H Awad}, \bibinfo{person}{Mostafa~Z
  Ali}, {and} \bibinfo{person}{Ponnuthurai~N Suganthan}.}
  \bibinfo{year}{2017}\natexlab{}.
\newblock \showarticletitle{Ensemble sinusoidal differential covariance matrix
  adaptation with Euclidean neighborhood for solving CEC2017 benchmark
  problems}. In \bibinfo{booktitle}{\emph{2017 IEEE Congress on Evolutionary
  Computation (CEC)}}. IEEE, \bibinfo{pages}{372--379}.
\newblock


\bibitem[\protect\citeauthoryear{Bajer, Pitra, Repick{\`y}, and
  Hole{\v{n}}a}{Bajer et~al\mbox{.}}{2019}]%
        {bajer2019gaussian}
\bibfield{author}{\bibinfo{person}{Luk{\'a}{\v{s}} Bajer},
  \bibinfo{person}{Zbyn{\v{e}}k Pitra}, \bibinfo{person}{Jakub Repick{\`y}},
  {and} \bibinfo{person}{Martin Hole{\v{n}}a}.}
  \bibinfo{year}{2019}\natexlab{}.
\newblock \showarticletitle{Gaussian process surrogate models for the CMA
  evolution strategy}.
\newblock \bibinfo{journal}{\emph{Evolutionary computation}}
  \bibinfo{volume}{27}, \bibinfo{number}{4} (\bibinfo{year}{2019}),
  \bibinfo{pages}{665--697}.
\newblock


\bibitem[\protect\citeauthoryear{Biswas, Saha, De, Cobb, Das, and
  Jalaian}{Biswas et~al\mbox{.}}{2021}]%
        {biswas2021improving}
\bibfield{author}{\bibinfo{person}{Subhodip Biswas}, \bibinfo{person}{Debanjan
  Saha}, \bibinfo{person}{Shuvodeep De}, \bibinfo{person}{Adam~D Cobb},
  \bibinfo{person}{Swagatam Das}, {and} \bibinfo{person}{Brian~A Jalaian}.}
  \bibinfo{year}{2021}\natexlab{}.
\newblock \showarticletitle{Improving differential evolution through bayesian
  hyperparameter optimization}. In \bibinfo{booktitle}{\emph{2021 IEEE Congress
  on Evolutionary Computation (CEC)}}. IEEE, \bibinfo{pages}{832--840}.
\newblock


\bibitem[\protect\citeauthoryear{Boussa{\"\i}d, Lepagnot, and
  Siarry}{Boussa{\"\i}d et~al\mbox{.}}{2013}]%
        {boussaid2013survey}
\bibfield{author}{\bibinfo{person}{Ilhem Boussa{\"\i}d},
  \bibinfo{person}{Julien Lepagnot}, {and} \bibinfo{person}{Patrick Siarry}.}
  \bibinfo{year}{2013}\natexlab{}.
\newblock \showarticletitle{A survey on optimization metaheuristics}.
\newblock \bibinfo{journal}{\emph{Information sciences}}  \bibinfo{volume}{237}
  (\bibinfo{year}{2013}), \bibinfo{pages}{82--117}.
\newblock


\bibitem[\protect\citeauthoryear{Brest, Greiner, Boskovic, Mernik, and
  Zumer}{Brest et~al\mbox{.}}{2006}]%
        {brest2006self}
\bibfield{author}{\bibinfo{person}{Janez Brest}, \bibinfo{person}{Sao Greiner},
  \bibinfo{person}{Borko Boskovic}, \bibinfo{person}{Marjan Mernik}, {and}
  \bibinfo{person}{Viljem Zumer}.} \bibinfo{year}{2006}\natexlab{}.
\newblock \showarticletitle{Self-adapting control parameters in differential
  evolution: A comparative study on numerical benchmark problems}.
\newblock \bibinfo{journal}{\emph{IEEE transactions on evolutionary
  computation}} \bibinfo{volume}{10}, \bibinfo{number}{6}
  (\bibinfo{year}{2006}), \bibinfo{pages}{646--657}.
\newblock


\bibitem[\protect\citeauthoryear{Brest, Mau{\v{c}}ec, and
  Bo{\v{s}}kovi{\'c}}{Brest et~al\mbox{.}}{2020}]%
        {brest2020differential}
\bibfield{author}{\bibinfo{person}{Janez Brest}, \bibinfo{person}{Mirjam~Sepesy
  Mau{\v{c}}ec}, {and} \bibinfo{person}{Borko Bo{\v{s}}kovi{\'c}}.}
  \bibinfo{year}{2020}\natexlab{}.
\newblock \showarticletitle{Differential evolution algorithm for single
  objective bound-constrained optimization: Algorithm j2020}. In
  \bibinfo{booktitle}{\emph{2020 IEEE Congress on Evolutionary Computation
  (CEC)}}. IEEE, \bibinfo{pages}{1--8}.
\newblock


\bibitem[\protect\citeauthoryear{Cressie}{Cressie}{1990}]%
        {cressie1990origins}
\bibfield{author}{\bibinfo{person}{Noel Cressie}.}
  \bibinfo{year}{1990}\natexlab{}.
\newblock \showarticletitle{The origins of kriging}.
\newblock \bibinfo{journal}{\emph{Mathematical geology}} \bibinfo{volume}{22},
  \bibinfo{number}{3} (\bibinfo{year}{1990}), \bibinfo{pages}{239--252}.
\newblock


\bibitem[\protect\citeauthoryear{Hansen}{Hansen}{2019}]%
        {hansen2019global}
\bibfield{author}{\bibinfo{person}{Nikolaus Hansen}.}
  \bibinfo{year}{2019}\natexlab{}.
\newblock \showarticletitle{A global surrogate assisted CMA-ES}. In
  \bibinfo{booktitle}{\emph{Proceedings of the Genetic and Evolutionary
  Computation Conference}}. \bibinfo{pages}{664--672}.
\newblock


\bibitem[\protect\citeauthoryear{Hansen, M{\"u}ller, and Koumoutsakos}{Hansen
  et~al\mbox{.}}{2003}]%
        {hansen2003reducing}
\bibfield{author}{\bibinfo{person}{Nikolaus Hansen}, \bibinfo{person}{Sibylle~D
  M{\"u}ller}, {and} \bibinfo{person}{Petros Koumoutsakos}.}
  \bibinfo{year}{2003}\natexlab{}.
\newblock \showarticletitle{Reducing the time complexity of the derandomized
  evolution strategy with covariance matrix adaptation (CMA-ES)}.
\newblock \bibinfo{journal}{\emph{Evolutionary computation}}
  \bibinfo{volume}{11}, \bibinfo{number}{1} (\bibinfo{year}{2003}),
  \bibinfo{pages}{1--18}.
\newblock


\bibitem[\protect\citeauthoryear{Helton and Davis}{Helton and Davis}{2003}]%
        {helton2003latin}
\bibfield{author}{\bibinfo{person}{Jon~C Helton} {and}
  \bibinfo{person}{Freddie~Joe Davis}.} \bibinfo{year}{2003}\natexlab{}.
\newblock \showarticletitle{Latin hypercube sampling and the propagation of
  uncertainty in analyses of complex systems}.
\newblock \bibinfo{journal}{\emph{Reliability Engineering \& System Safety}}
  \bibinfo{volume}{81}, \bibinfo{number}{1} (\bibinfo{year}{2003}),
  \bibinfo{pages}{23--69}.
\newblock


\bibitem[\protect\citeauthoryear{Jin}{Jin}{2011}]%
        {jin2011surrogate}
\bibfield{author}{\bibinfo{person}{Yaochu Jin}.}
  \bibinfo{year}{2011}\natexlab{}.
\newblock \showarticletitle{Surrogate-assisted evolutionary computation: Recent
  advances and future challenges}.
\newblock \bibinfo{journal}{\emph{Swarm and Evolutionary Computation}}
  \bibinfo{volume}{1}, \bibinfo{number}{2} (\bibinfo{year}{2011}),
  \bibinfo{pages}{61--70}.
\newblock


\bibitem[\protect\citeauthoryear{Jones, Schonlau, and Welch}{Jones
  et~al\mbox{.}}{1998}]%
        {jones1998efficient}
\bibfield{author}{\bibinfo{person}{Donald~R Jones}, \bibinfo{person}{Matthias
  Schonlau}, {and} \bibinfo{person}{William~J Welch}.}
  \bibinfo{year}{1998}\natexlab{}.
\newblock \showarticletitle{Efficient global optimization of expensive
  black-box functions}.
\newblock \bibinfo{journal}{\emph{Journal of Global optimization}}
  \bibinfo{volume}{13}, \bibinfo{number}{4} (\bibinfo{year}{1998}),
  \bibinfo{pages}{455--492}.
\newblock


\bibitem[\protect\citeauthoryear{Kendall}{Kendall}{1938}]%
        {kendall1938new}
\bibfield{author}{\bibinfo{person}{Maurice~G Kendall}.}
  \bibinfo{year}{1938}\natexlab{}.
\newblock \showarticletitle{A new measure of rank correlation}.
\newblock \bibinfo{journal}{\emph{Biometrika}} \bibinfo{volume}{30},
  \bibinfo{number}{1/2} (\bibinfo{year}{1938}), \bibinfo{pages}{81--93}.
\newblock


\bibitem[\protect\citeauthoryear{Kern, Hansen, and Koumoutsakos}{Kern
  et~al\mbox{.}}{2006}]%
        {kern2006local}
\bibfield{author}{\bibinfo{person}{Stefan Kern}, \bibinfo{person}{Nikolaus
  Hansen}, {and} \bibinfo{person}{Petros Koumoutsakos}.}
  \bibinfo{year}{2006}\natexlab{}.
\newblock \showarticletitle{Local meta-models for optimization using evolution
  strategies}.
\newblock In \bibinfo{booktitle}{\emph{Parallel Problem Solving from
  Nature-PPSN IX}}. \bibinfo{publisher}{Springer}, \bibinfo{pages}{939--948}.
\newblock


\bibitem[\protect\citeauthoryear{Mohamed, Hadi, Mohamed, Agrawal, Kumar, and
  Suganthan}{Mohamed et~al\mbox{.}}{[n.\,d.]}]%
        {cec2021}
\bibfield{author}{\bibinfo{person}{Ali~Wagdy Mohamed}, \bibinfo{person}{Anas~A
  Hadi}, \bibinfo{person}{Ali~Khater Mohamed}, \bibinfo{person}{Prachi
  Agrawal}, \bibinfo{person}{Abhishek Kumar}, {and} \bibinfo{person}{P.N
  Suganthan}.} \bibinfo{year}{[n.\,d.]}\natexlab{}.
\newblock \bibinfo{title}{Problem Definitions and Evaluation Criteria for the
  CEC 2021 Special Session and Competition on Single Objective Bound
  Constrained Numerical Optimization}.
\newblock
\newblock
\newblock
\shownote{\url{https://github.com/P-N-Suganthan/2021-SO-BCO/blob/main/CEC2021\%20TR_final\%20(1).pdf}}.


\bibitem[\protect\citeauthoryear{Mohamed, Hadi, Mohamed, and Awad}{Mohamed
  et~al\mbox{.}}{2020}]%
        {mohamed2020evaluating}
\bibfield{author}{\bibinfo{person}{Ali~Wagdy Mohamed}, \bibinfo{person}{Anas~A
  Hadi}, \bibinfo{person}{Ali~Khater Mohamed}, {and} \bibinfo{person}{Noor~H
  Awad}.} \bibinfo{year}{2020}\natexlab{}.
\newblock \showarticletitle{Evaluating the performance of adaptive
  GainingSharing knowledge based algorithm on CEC 2020 benchmark problems}. In
  \bibinfo{booktitle}{\emph{2020 IEEE Congress on Evolutionary Computation
  (CEC)}}. IEEE, \bibinfo{pages}{1--8}.
\newblock


\bibitem[\protect\citeauthoryear{Nishida and Akimoto}{Nishida and
  Akimoto}{2018}]%
        {nishida2018benchmarking}
\bibfield{author}{\bibinfo{person}{Kouhei Nishida} {and}
  \bibinfo{person}{Youhei Akimoto}.} \bibinfo{year}{2018}\natexlab{}.
\newblock \showarticletitle{Benchmarking the PSA-CMA-ES on the BBOB noiseless
  testbed}. In \bibinfo{booktitle}{\emph{Proceedings of the Genetic and
  Evolutionary Computation Conference Companion}}. \bibinfo{pages}{1529--1536}.
\newblock


\bibitem[\protect\citeauthoryear{Okulewicz and Zaborski}{Okulewicz and
  Zaborski}{2021}]%
        {okulewicz2021benchmarking}
\bibfield{author}{\bibinfo{person}{Micha{\l} Okulewicz} {and}
  \bibinfo{person}{Mateusz Zaborski}.} \bibinfo{year}{2021}\natexlab{}.
\newblock \showarticletitle{Benchmarking SHADE algorithm enhanced with model
  based optimization on the BBOB noiseless testbed}. In
  \bibinfo{booktitle}{\emph{Proceedings of the Genetic and Evolutionary
  Computation Conference Companion}}. \bibinfo{pages}{1259--1266}.
\newblock


\bibitem[\protect\citeauthoryear{Qin, Huang, and Suganthan}{Qin
  et~al\mbox{.}}{2008}]%
        {qin2008differential}
\bibfield{author}{\bibinfo{person}{A~Kai Qin}, \bibinfo{person}{Vicky~Ling
  Huang}, {and} \bibinfo{person}{Ponnuthurai~N Suganthan}.}
  \bibinfo{year}{2008}\natexlab{}.
\newblock \showarticletitle{Differential evolution algorithm with strategy
  adaptation for global numerical optimization}.
\newblock \bibinfo{journal}{\emph{IEEE transactions on Evolutionary
  Computation}} \bibinfo{volume}{13}, \bibinfo{number}{2}
  (\bibinfo{year}{2008}), \bibinfo{pages}{398--417}.
\newblock


\bibitem[\protect\citeauthoryear{Sallam, Elsayed, Chakrabortty, and
  Ryan}{Sallam et~al\mbox{.}}{2020}]%
        {sallam2020improved}
\bibfield{author}{\bibinfo{person}{Karam~M Sallam}, \bibinfo{person}{Saber~M
  Elsayed}, \bibinfo{person}{Ripon~K Chakrabortty}, {and}
  \bibinfo{person}{Michael~J Ryan}.} \bibinfo{year}{2020}\natexlab{}.
\newblock \showarticletitle{Improved multi-operator differential evolution
  algorithm for solving unconstrained problems}. In
  \bibinfo{booktitle}{\emph{2020 IEEE Congress on Evolutionary Computation
  (CEC)}}. IEEE, \bibinfo{pages}{1--8}.
\newblock


\bibitem[\protect\citeauthoryear{Storn and Price}{Storn and Price}{1997}]%
        {storn1997differential}
\bibfield{author}{\bibinfo{person}{Rainer Storn} {and} \bibinfo{person}{Kenneth
  Price}.} \bibinfo{year}{1997}\natexlab{}.
\newblock \showarticletitle{Differential evolution--a simple and efficient
  heuristic for global optimization over continuous spaces}.
\newblock \bibinfo{journal}{\emph{Journal of global optimization}}
  \bibinfo{volume}{11}, \bibinfo{number}{4} (\bibinfo{year}{1997}),
  \bibinfo{pages}{341--359}.
\newblock


\bibitem[\protect\citeauthoryear{Suganthan}{Suganthan}{[n.\,d.]}]%
        {codeoftopmethods}
\bibfield{author}{\bibinfo{person}{P.N Suganthan}.}
  \bibinfo{year}{[n.\,d.]}\natexlab{}.
\newblock \bibinfo{title}{Code of top methods}.
\newblock
\newblock
\newblock
\shownote{\url{https://github.com/P-N-Suganthan/2021-SO-BCO/blob/main/Codes-of-top-methods\%20(1).zip}}.


\bibitem[\protect\citeauthoryear{Tanabe and Fukunaga}{Tanabe and
  Fukunaga}{2013}]%
        {tanabe2013success}
\bibfield{author}{\bibinfo{person}{Ryoji Tanabe} {and} \bibinfo{person}{Alex
  Fukunaga}.} \bibinfo{year}{2013}\natexlab{}.
\newblock \showarticletitle{Success-history based parameter adaptation for
  differential evolution}. In \bibinfo{booktitle}{\emph{2013 IEEE congress on
  evolutionary computation}}. IEEE, \bibinfo{pages}{71--78}.
\newblock


\bibitem[\protect\citeauthoryear{Tanabe and Fukunaga}{Tanabe and
  Fukunaga}{2015}]%
        {tanabe2015tuning}
\bibfield{author}{\bibinfo{person}{Ryoji Tanabe} {and} \bibinfo{person}{Alex
  Fukunaga}.} \bibinfo{year}{2015}\natexlab{}.
\newblock \showarticletitle{Tuning differential evolution for cheap, medium,
  and expensive computational budgets}. In \bibinfo{booktitle}{\emph{2015 IEEE
  Congress on Evolutionary Computation (CEC)}}. IEEE,
  \bibinfo{pages}{2018--2025}.
\newblock


\bibitem[\protect\citeauthoryear{Tanabe and Fukunaga}{Tanabe and
  Fukunaga}{2014}]%
        {tanabe2014improving}
\bibfield{author}{\bibinfo{person}{Ryoji Tanabe} {and} \bibinfo{person}{Alex~S
  Fukunaga}.} \bibinfo{year}{2014}\natexlab{}.
\newblock \showarticletitle{Improving the search performance of SHADE using
  linear population size reduction}. In \bibinfo{booktitle}{\emph{2014 IEEE
  congress on evolutionary computation (CEC)}}. IEEE,
  \bibinfo{pages}{1658--1665}.
\newblock


\bibitem[\protect\citeauthoryear{Vikhar}{Vikhar}{2016}]%
        {vikhar2016evolutionary}
\bibfield{author}{\bibinfo{person}{Pradnya~A Vikhar}.}
  \bibinfo{year}{2016}\natexlab{}.
\newblock \showarticletitle{Evolutionary algorithms: A critical review and its
  future prospects}. In \bibinfo{booktitle}{\emph{2016 International conference
  on global trends in signal processing, information computing and
  communication (ICGTSPICC)}}. IEEE, \bibinfo{pages}{261--265}.
\newblock


\bibitem[\protect\citeauthoryear{Weisberg}{Weisberg}{2013}]%
        {weisberg2013applied}
\bibfield{author}{\bibinfo{person}{Sanford Weisberg}.}
  \bibinfo{year}{2013}\natexlab{}.
\newblock \bibinfo{booktitle}{\emph{Applied linear regression}}.
\newblock \bibinfo{publisher}{John Wiley \& Sons}.
\newblock


\bibitem[\protect\citeauthoryear{Yamaguchi and Akimoto}{Yamaguchi and
  Akimoto}{2017}]%
        {yamaguchi2017benchmarking}
\bibfield{author}{\bibinfo{person}{Takahiro Yamaguchi} {and}
  \bibinfo{person}{Youhei Akimoto}.} \bibinfo{year}{2017}\natexlab{}.
\newblock \showarticletitle{Benchmarking the novel CMA-ES restart strategy
  using the search history on the BBOB noiseless testbed}. In
  \bibinfo{booktitle}{\emph{Proceedings of the Genetic and Evolutionary
  Computation Conference Companion}}. \bibinfo{pages}{1780--1787}.
\newblock


\bibitem[\protect\citeauthoryear{Zaborski, Okulewicz, and
  Ma{\'n}dziuk}{Zaborski et~al\mbox{.}}{2020}]%
        {zaborski2020analysis}
\bibfield{author}{\bibinfo{person}{Mateusz Zaborski},
  \bibinfo{person}{Micha{\l} Okulewicz}, {and} \bibinfo{person}{Jacek
  Ma{\'n}dziuk}.} \bibinfo{year}{2020}\natexlab{}.
\newblock \showarticletitle{Analysis of statistical model-based optimization
  enhancements in Generalized Self-Adapting Particle Swarm Optimization
  framework}.
\newblock \bibinfo{journal}{\emph{Foundations of Computing and Decision
  Sciences}} \bibinfo{volume}{45}, \bibinfo{number}{3} (\bibinfo{year}{2020}),
  \bibinfo{pages}{233--254}.
\newblock


\bibitem[\protect\citeauthoryear{Zhang and Sanderson}{Zhang and
  Sanderson}{2009}]%
        {zhang2009jade}
\bibfield{author}{\bibinfo{person}{Jingqiao Zhang} {and}
  \bibinfo{person}{Arthur~C Sanderson}.} \bibinfo{year}{2009}\natexlab{}.
\newblock \showarticletitle{JADE: adaptive differential evolution with optional
  external archive}.
\newblock \bibinfo{journal}{\emph{IEEE Transactions on evolutionary
  computation}} \bibinfo{volume}{13}, \bibinfo{number}{5}
  (\bibinfo{year}{2009}), \bibinfo{pages}{945--958}.
\newblock


\end{thebibliography}

\onecolumn

\newpage

\appendix

\section*{\textbf{--- \textit{Supplementary material} --- }}

\vspace{0.3cm}

\begin{figure*}[h]
	\begin{center}
		\includegraphics[width=1.0\textwidth]%
		{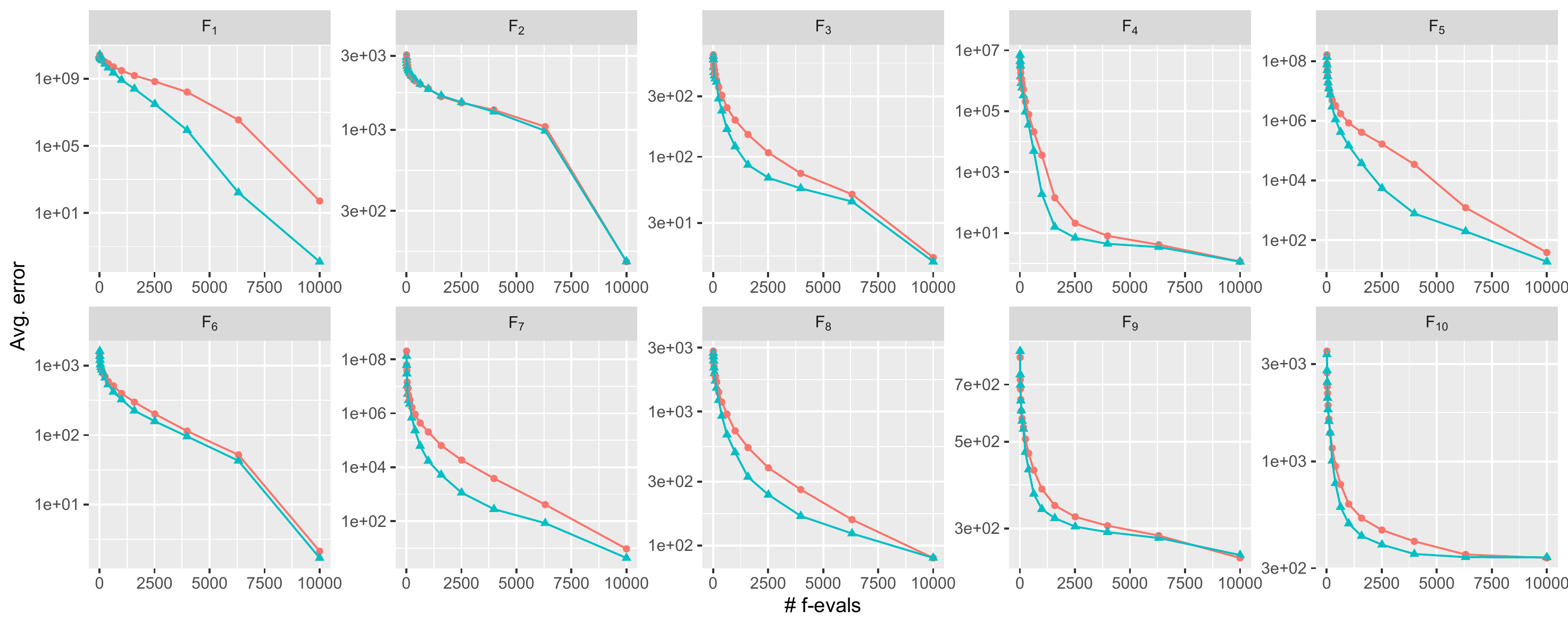}
	\end{center}
    \vspace*{-5mm}
	\caption{The averaged convergence of psLSHADE (blue line) and LSHADE (red line) for all 10D functions from CEC2021 with $10^3 \cdot D$ optimization budget. The x-axis represents the number of FFEs and the y-axis the average error.
	\label{fig:convergence10}}
\end{figure*}

In this supplementary material we extend the results presented in the main body of the paper. 
Figure~\ref{fig:convergence10}, for each 10D function from CEC2021~\cite{cec2021}, illustrates the convergence of psLSHADE and LSHADE calculated as the average error value after each FFE count point. This figure is an analog of Figure~1 in the main text which shows the same plots for 20D functions.
%Figure~\ref{fig:convergence10} refers to 10D functions, while Figure~\ref{fig:convergence20} to 20D functions. Only Figure~\ref{fig:convergence20} is presented in the source work and refers to the original Figure 1. 
Generally speaking, the result for 10D and 20D are similar when comparing the individual functions and the conclusions drawn in the paper with respect to 20D functions are also valid for 10D.

\begin{figure*}[h]
	\begin{center}
		\includegraphics[width=1.0\textwidth]%
		{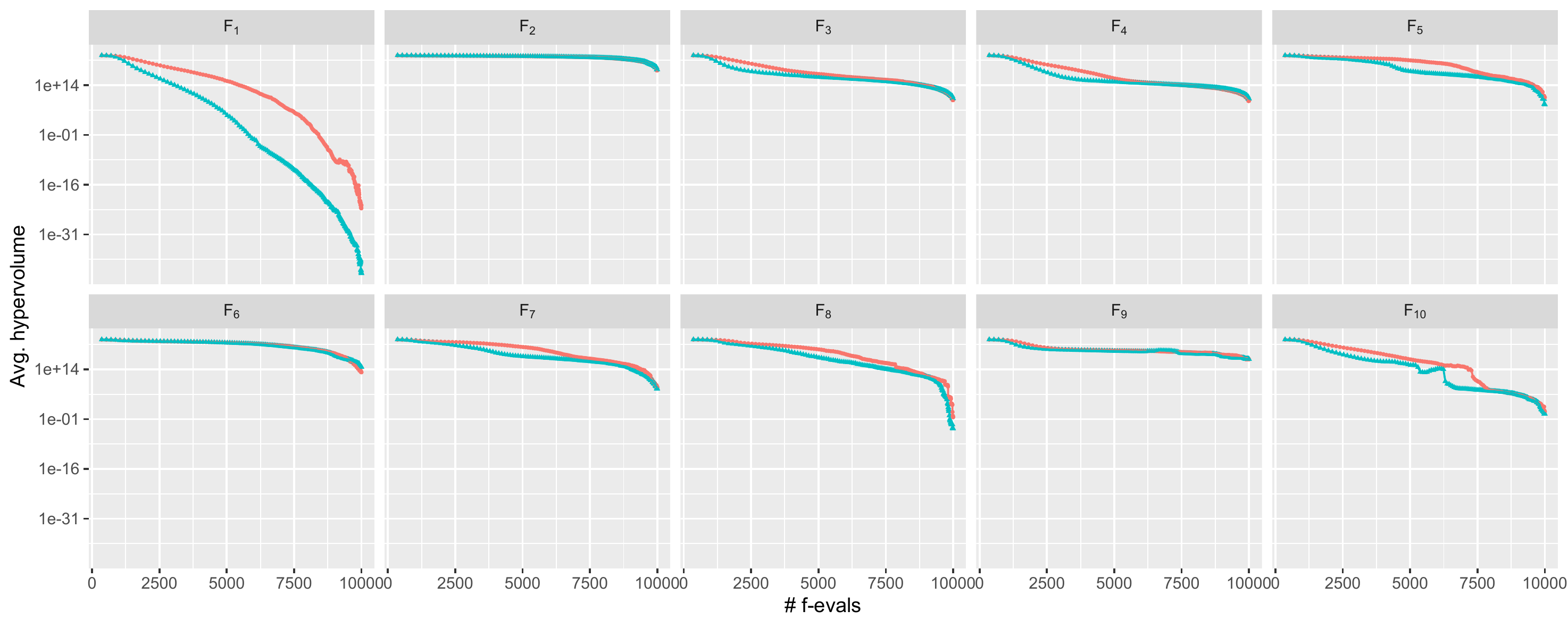}
	\end{center}
    \vspace*{-5mm}
		\caption{The hyper-volme of psLSHADE (blue line) and LSHADE (red line) for all 10D functions from CEC2021 with $10^3 \cdot D$ optimization budget. The x-axis represents the number of FFEs and the y-axis the average hypervolume size in each generation.
	\label{fig:hypervolume_10}}
\end{figure*}

\begin{figure*}[h]
	\begin{center}
		\includegraphics[width=1.0\textwidth]%
		{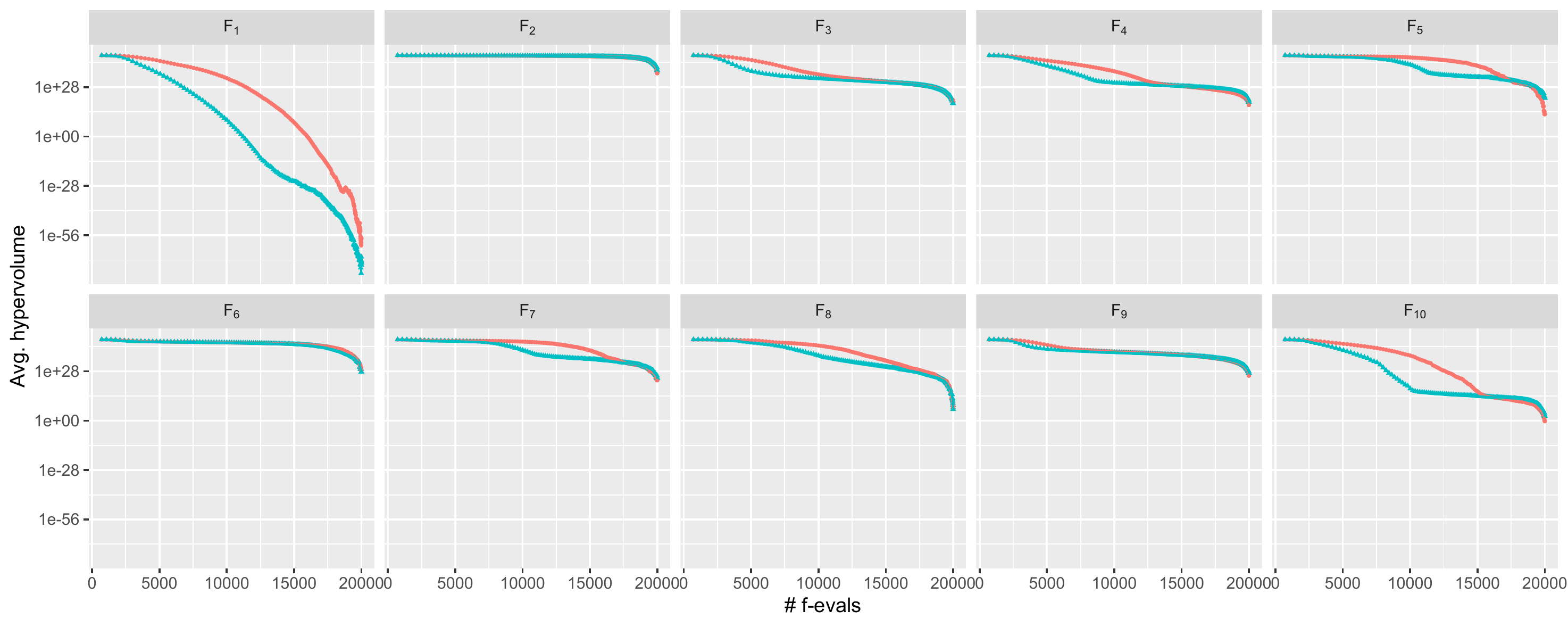}
	\end{center}
    \vspace*{-5mm}
		\caption{The hyper-volme of psLSHADE (blue line) and LSHADE (red line) for all 20D functions from CEC2021 with $10^3 \cdot D$ optimization budget. The x-axis represents the number of FFEs and the y-axis the average hypervolume size in each generation.
	\label{fig:hypervolume_20}}
\end{figure*}

Figures~\ref{fig:hypervolume_10} and~\ref{fig:hypervolume_20} present the changes of the hyper-volumes of psLSHADE and LSHADE during an optimization run, for 10D and 20D functions, respectively.
%Figure~\ref{fig:hypervolume_10} refers to 10D functions, while Figure~\ref{fig:hypervolume_20}  to 20D functions. 
In the main body of the paper (Figure~3), hyper-volume changes are demonstrated only for $F_1$ and $F_2$ in 20D.
The impact of pre-screening on the hyper-volume is visible for all functions in both dimensions (10D and 20D), however its scale varies depending on the function. 

\begin{figure*}[h]
	\begin{center}
		\includegraphics[width=1.0\textwidth]%
		{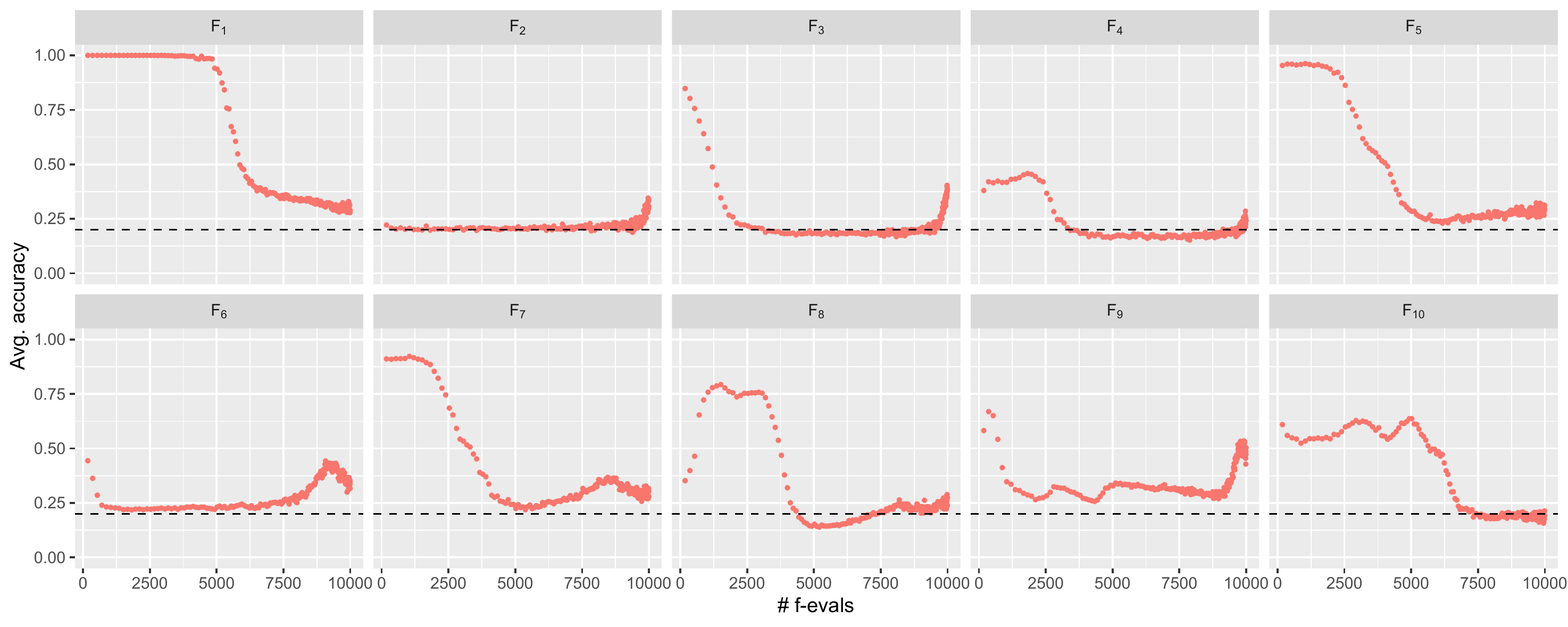}
	\end{center}
    \vspace*{-5mm}
		\caption{The averaged accuracy of meta-model selection in each generation for all 10D functions from CEC2021 with $10^3 \cdot D$ optimization budget. The x-axis represents the number of FFEs and the y-axis the average accuracy.
	\label{fig:accuracy_10}}
\end{figure*}

\begin{figure*}[h]
	\begin{center}
		\includegraphics[width=1.0\textwidth]%
		{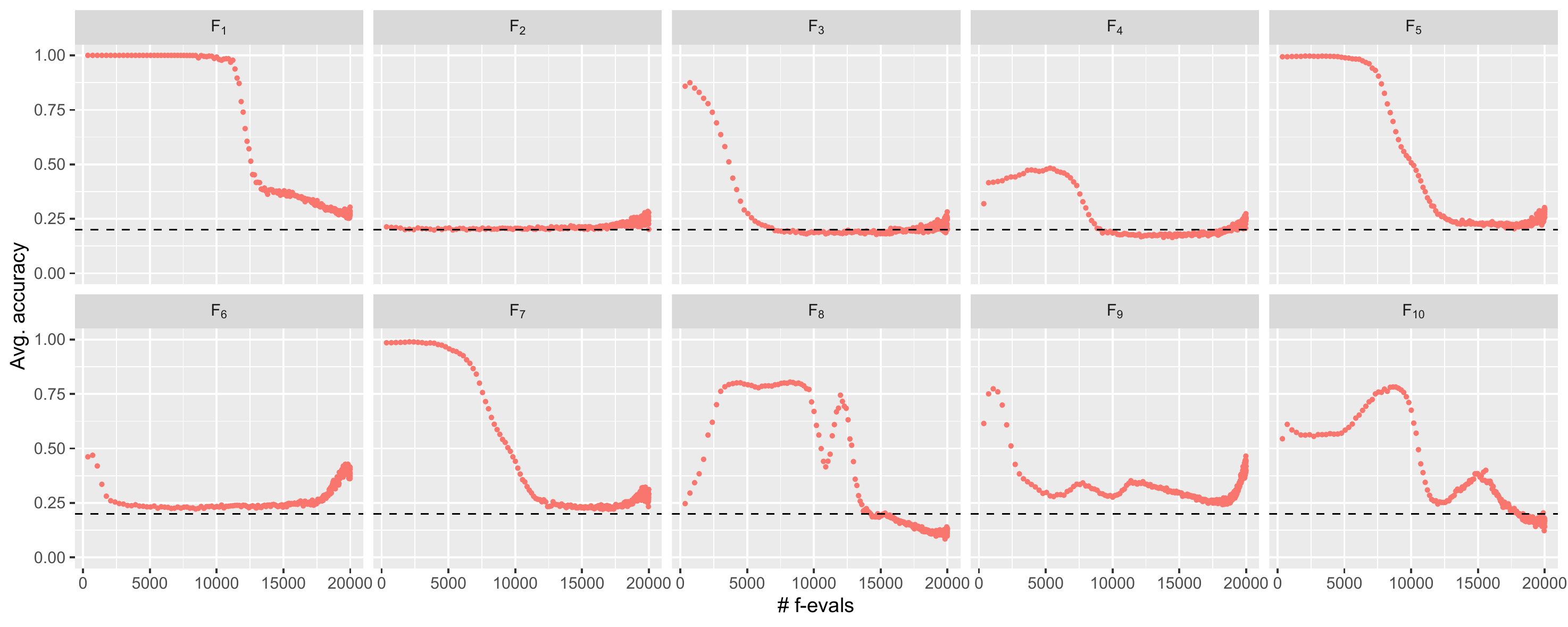}
	\end{center}
    \vspace*{-5mm}
		\caption{The average accuracy of meta-model selection in each generation for all 20D functions from CEC2021 with $10^3 \cdot D$ optimization budget. The x-axis represents the number of FFEs and the y-axis the average accuracy.
	\label{fig:accuracy_20}}
\end{figure*}

Figures~\ref{fig:accuracy_10} and~\ref{fig:accuracy_20} present the average accuracy of the meta-model selections in each generation, for 10D and 20D functions, respectively. The figures extend Figure~4 from the main paper which presents the same results for 20D versions of the selected 4 functions: $F_1$, $F_2$, $F_8$, and $F_9$. The general trend of diminishing accuracy of estimates is apparent for all functions and does not vary significantly between 10D and 20D cases.  
The only exception is $F_8$, for which the characteristic for 20D is slightly different than for 10D (one can observe a spike in accuracy around $1.2 \cdot 10^4$ FFEs in 20D). A deeper analysis showed that this spike can be observed in all 4 transformations containing \textit{shift}. The 2D versions of $F_8$ and its contour plot are presented in Figure~\ref{fig:f8}. As can be seen in the figure, $F_8$ is a composition multi-modal function, but with a significant smooth area. This smooth area can be relatively well approximated by the meta-model. We hypothesize that the spike occurs when the archive begins to contain mainly samples belonging to the smooth area, which increases the accuracy. 
This phenomenon is not visible for the 10D case, most likely because the shift transformation in 10D is less severe due to the specific location of the global optimum.
A full explanation of this phenomenon is the subject of future research.
%Besides, only one combination of shift transformation is applied in the benchmark, so generalizing is challenging. 

\begin{figure*}[h]
	\begin{center}
		\includegraphics[width=1.0\textwidth]%
		{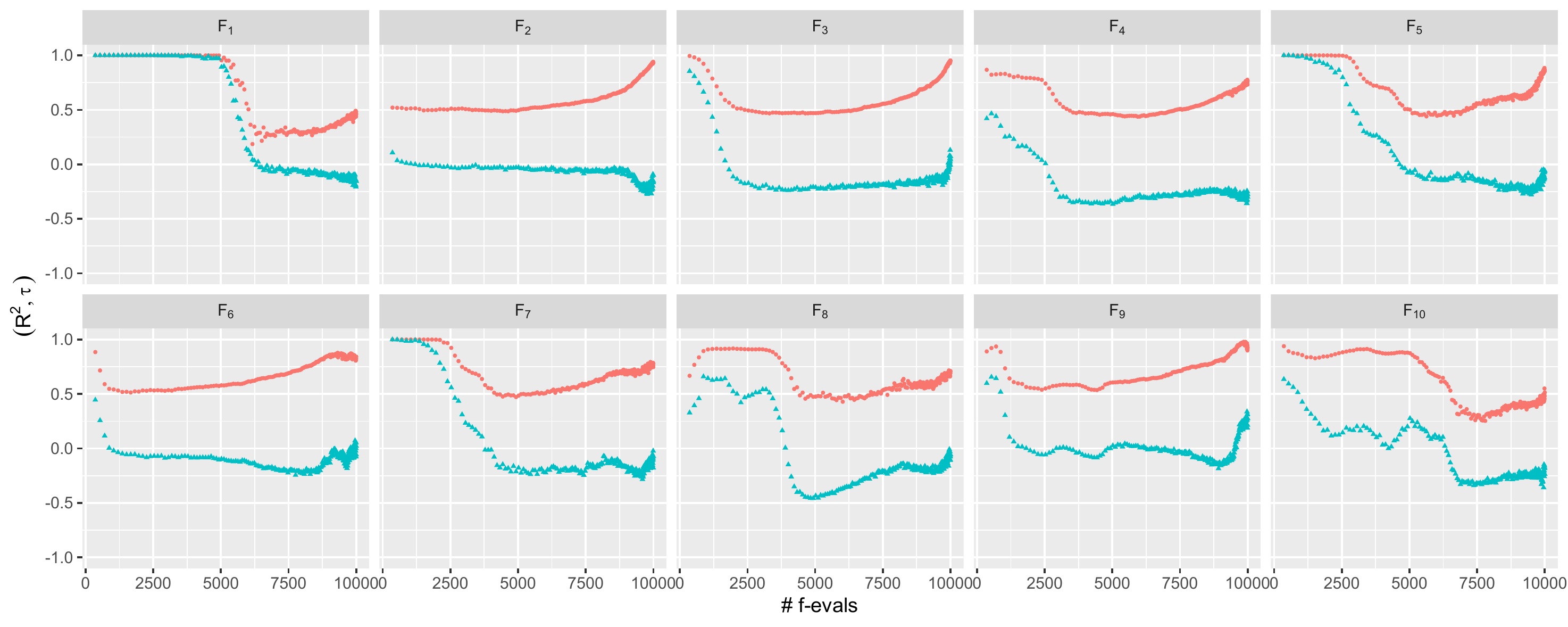}
	\end{center}
    \vspace*{-5mm}
	\caption{The average $R^2$ and Kendall's $\tau$ in each generation for all 10D functions from CEC2021 with $10^3 \cdot D$ optimization budget. The x-axis represents the number of FFEs and the y-axis the average values of $R^2$ and Kendall's $\tau$.
	\label{fig:r2_tau_10}}
\end{figure*}

\begin{figure*}[h]
	\begin{center}
		\includegraphics[width=1.0\textwidth]%
		{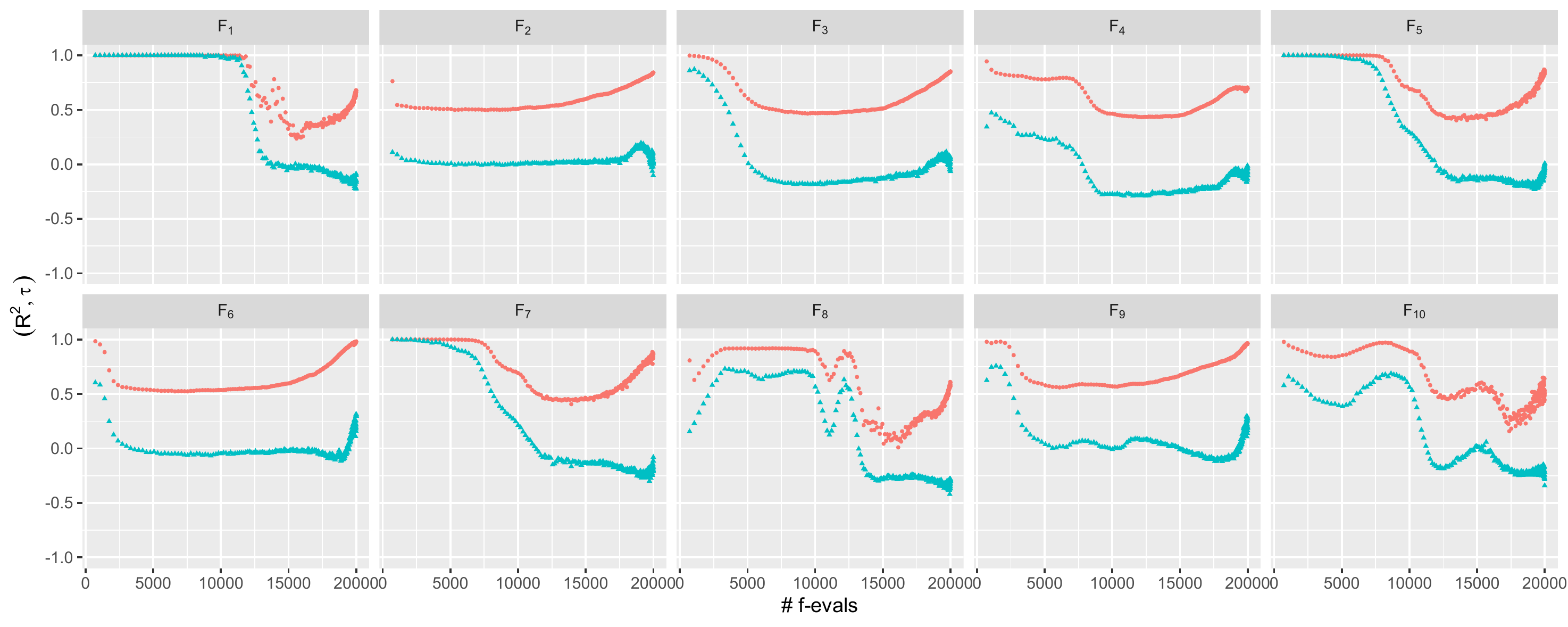}
	\end{center}
    \vspace*{-5mm}
	\caption{The average $R^2$ and Kendall's $\tau$ in each generation for all 20D functions from CEC2021 with $10^3 \cdot D$ optimization budget. The x-axis represents the number of FFEs and the y-axis the average values of $R^2$ and Kendall's $\tau$.
	\label{fig:r2_tau_20}}
\end{figure*}

\begin{figure*}[h]%
    \centering
    \subfloat[\centering The 3D maps of $2D$ version of composition function $F_8$]{{\includegraphics[width=7.0cm]{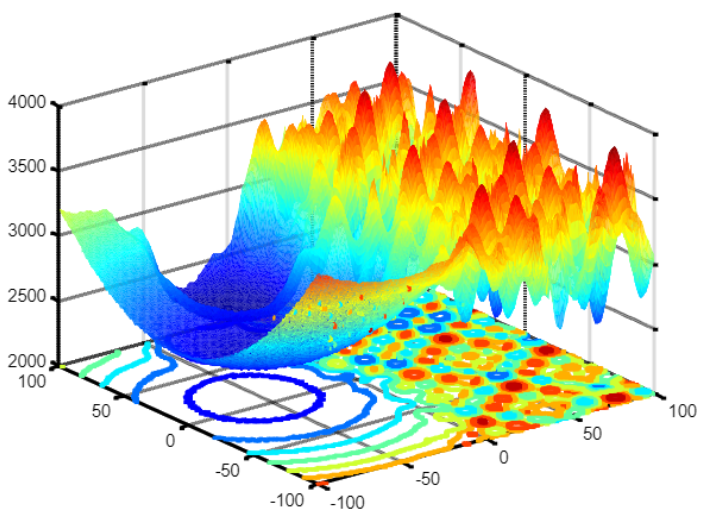} }}%
    \qquad
    \subfloat[\centering The 2D contour map of $2D$ version of composition function $F_8$]{{\includegraphics[width=7.0cm]{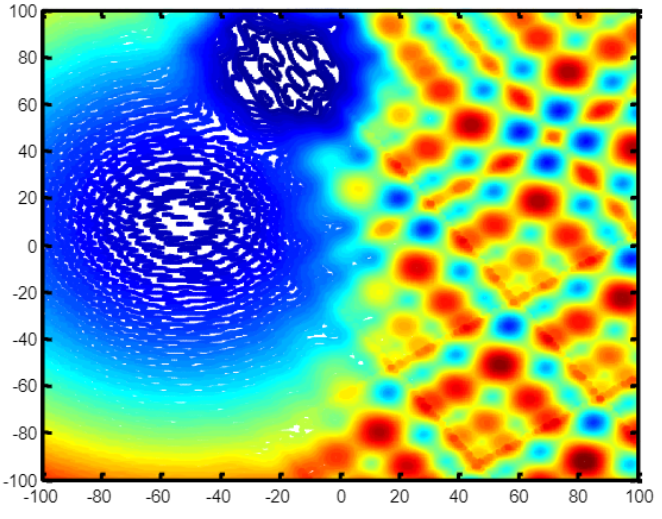} }}%
    \caption{The 3D maps of $2D$ version of composition function $F_1$ (a) and its contour map (b) The figures are reprinted from~\cite{cec2021}.}
    \label{fig:f8}%
\end{figure*}

Figures~\ref{fig:r2_tau_10} and~\ref{fig:r2_tau_20} depict the average $R^2$ and Kendall's $\tau$ for functions in 10D and 20D, respectively. The results extend Figure~5 in the main paper that relates exclusively to 20D versions of functions $F_1$, $F_2$, $F_8$ and $F_9$. 
%{\color{red}\st{The observations for the selected four functions at the source work are general enough to be applied to other functions.}} 
Generally, there are no particular differences between the 10D and 20D plots. although function $F_8$ again behaves a little differently between 10D and 20D cases. The reasons for this phenomenon are most probably the same as for the case of accuracy. The unusual increase of the meta-model fit, defined by $R^2$, further confirms the above-presented hypothesis.
 
% \begin{figure*}[h]
% 	\begin{center}
% 		\includegraphics[width=1.0\textwidth]%
% 		{figures/convergence_fun_20}
% 	\end{center}
%     \vspace*{-5mm}
% 	\caption{The averaged convergence of psLSHADE (blue line) and LSHADE (red line) for all 20D functions from CEC2021 with $10^3 \cdot D$ optimization budget. The x-axis represents the number of FFEs and the y-axis the average error (a difference with the optimum).
% 	\label{fig:convergence20}}
% \end{figure*}

\end{document}